\documentclass[10pt,a4paper,oneside]{article}
\usepackage[top=1in,bottom=1in,left=0.75in,right=0.75in]{geometry}
\usepackage{amsmath}
\usepackage{graphicx}
\usepackage{subfigure}
\usepackage{amssymb}
\numberwithin{equation}{section}
\usepackage{subfigure}
\usepackage{threeparttablex}
\usepackage{csquotes}
\usepackage{enumitem}
\usepackage{blindtext}
\usepackage{scrextend}
\addtokomafont{labelinglabel}{\sffamily}
\usepackage{epstopdf}
\usepackage{caption}
\usepackage{array}
\usepackage{lineno}
\usepackage{color,colortbl}
\usepackage{hyperref}
\hypersetup{
    colorlinks=true,
    linkcolor=blue,
    filecolor=magenta,      
    urlcolor=cyan,
}
\newcolumntype{Y}{>{\centering\arraybackslash}X}

\definecolor{Gray}{gray}{0.9}           
\definecolor{LightCyan}{rgb}{0.88,1,1}

\graphicspath{{Images/}}
\usepackage{cite}
\usepackage{epsfig}
\usepackage{amsmath,amssymb, amsthm,latexsym,graphics,graphicx,color}
\usepackage{booktabs}
\usepackage{algorithmic}
\usepackage{algorithm}
\usepackage{multirow}

\let\tempcommand\description
\renewcommand{\description}{
  \tempcommand
  \setlength{\itemsep}{1pt}
  \setlength{\parskip}{0pt}
  \setlength{\parsep}{0pt}
}

\graphicspath{{images/}}

\begin{document}
\title{Tensor Based Second Order Variational Model for \\ Image Reconstruction}
\author{Jinming Duan$^\dagger$, Wil OC Ward$^\dagger$, Luke Sibbett$^\dagger$, Zhenkuan Pan$^\ddagger$, Li Bai$^\dagger$\\
$^\dagger$School of Computer Science, University of Nottingham, UK \\
$^\ddagger$School of Computer Science and Technology, Qindao University, China}
\date{}
\maketitle

\begin{abstract}
\noindent Second order total variation (SOTV) models have advantages for image reconstruction over their first order counterparts including their ability to remove the staircase artefact in the reconstructed image, but they tend to blur the reconstructed image. To overcome this drawback, we introduce a new Tensor Weighted Second Order (TWSO) model for image reconstruction. Specifically, we develop a novel regulariser for the SOTV model that uses the Frobenius norm of the product of the SOTV Hessian matrix and the anisotropic tensor. We then adapt the alternating direction method of multipliers (ADMM) to solve the proposed model by breaking down the original problem into several subproblems. All the subproblems have closed-forms and can thus be solved efficiently. The proposed method is compared with a range of state-of-the-art approaches such as tensor-based anisotropic diffusion, total generalised variation, Euler's elastica, etc. Numerical experimental results of the method on both synthetic and real images from the Berkeley database BSDS500 demonstrate that the proposed method eliminates both the staircase and blurring effects and outperforms the existing approaches for image inpainting and denoising applications.\\

\end{abstract}

\section{Introduction}
Nonlinear diffusion filters have proven to be effective for multiscale image processing in the past two decades. The most influential nonlinear image diffusion model was proposed by Perona and Malik (PM) \cite{perona1990scale} 
\begin{equation}
{\partial _t}u = div\left( {g\left( {\left| {\nabla {u }} \right|} \right)\nabla u} \right), \label{eq:ISO}
\end{equation}
where $g\left( {\left| {\nabla {u }} \right|} \right)$ is a non-increasing edge diffusivity function. The PM formulation can produce visually pleasing image denoising results and preserve the image structure. However, noise cannot be removed at the edges/lines of the image. Weickert \cite{weickert1998anisotropic} thus replaced the edge diffusivity function with an anisotropic diffusion tensor $\textbf{T}$ and proposed the following tensor-based diffusion formulation
\begin{equation}
{\partial _t}u = div\left( {\textbf{T}\nabla u} \right). \label{eq:CED}
\end{equation}
This filter can reduce noise along edges and simultaneously preserve image structure. Compared with the pure edge-based descriptors, tensor can measure the image geometry in the neighbourhood of each image pixel, thus offering a richer description for an image. This makes it more appealing to use in various image processing techniques.

Technically, an image diffusion model is a partial differential equation (PDE), and minimising an energy functional of an image with the variational framework normally results in a PDE. Various image diffusion models can be
thereby derived directly from the variational framework. For instance, if $g\left( {\left| {\nabla u} \right|} \right) = {1 \mathord{\left/{\vphantom {1 {\left| {\nabla u} \right|}}} \right.\kern-\nulldelimiterspace} {\left| {\nabla u} \right|}}$, the PM model (\ref{eq:ISO}) can be derived from the total variation energy functional $\int_\Omega  {\left| {\nabla u} \right|} dx$ using the gradient descent optimisation method. The variational approach has several advantages over the PDE/diffusion-based method \cite{roussos2010tensor}, including easy integration of constraints imposed by the problems and employing powerful modern optimisation techniques such as primal-dual \cite{chambolle2011first}, fast iterative shrinkage-thresholding algorithm \cite{beck2009fast}, and alternating direction method of multipliers \cite{boyd2011distributed}. 

Several works have investigated the tensor-based variational approach. Krajsek and Scharr \cite{krajsek2010diffusion} developed a linear anisotropic regularisation term that forms the basis of a tensor-valued energy functional for image denoising. Grasmair and Lenzen \cite{grasmair2010anisotropic,lenzen2013class} penalised image variation by introducing a diffusion tensor that depends on the structure tensor of the image. Roussous and Maragos \cite{roussos2010tensor} considered a functional that utilises only the eigenvalues of the structure tensor. Freddie et al. proposed an extended anisotropic diffusion model \cite{aastrom2012tensor}, which is a tensor-based variational formulation, for colour image denoising. They discarded the Gaussian convolution when computing the structure tensor. Hence, the Euler-equation of the functional can be elegantly derived. They later introduced a tensor-based functional named the gradient energy total variation \cite{aastrom2015tensor}, which allows them to consider both eigenvalues and eigenvectors of the gradient energy tensor. Furthermore, Lefkimmiatis et al. \cite{lefkimmiatis2015structure} considered the Schatten-norm of the structure tensor eigenvalues, which is similar to Roussous and Maragos' work \cite{roussos2010tensor}.

However, these existing works mentioned above only consider the standard first order total variation (FOTV) energy. A drawback of the FOTV model is that it favours piecewise-constant solutions. Thus, it can create strong staircase artefacts in the smooth regions of the restored image. Another drawback of the FOTV model is its use of gradient magnitude to penalise image variations at pixel locations $\textbf{x}$ that relies only on $\textbf{x}$ without taking into account the pixel neighbourhood. As such, the FOTV has difficulties inpainting images with large gaps. High order variational models have been widely applied to remedy these side effects. Among these is the second order total variation (SOTV) model \cite{lysaker2003noise,bergounioux2010second,papafitsoros2013combined,papafitsoros2014combined,lefkimmiatis2013hessian}. Unlike the high order variational models, such as the Gaussian curvature \cite{brito2015image}, mean curvature \cite{zhu2013augmented}, Euler's elastica \cite{tai2011fast} etc., the SOTV is a convex high order extension of the FOTV, which guarantees a global solution. The SOTV is also more efficient to implement \cite{duan2016edge} than the total generalised variation (TGV) \cite{bredies2010total}. However, the inpainting results of the model highly depend on the geometry of the inpainting region, and it also tends to blur the inpainted area \cite{papafitsoros2013combined,papafitsoros2014combined}. For image denoising, the model also tends to blur object edges due to the fact that it imposes too much regularity on the image.

In this paper we propose a tensor-weighted second order (TWSO) variational model for image inpainting and denoising. A novel regulariser has been developed for the TWSO that uses the Frobenius norm of the product of the SOTV Hessian matrix and an anisotropic tensor, so that the TWSO can eliminate the staircase effect whilst preserve the sharp edges/boundaries of objects in the denoised image. For image inpainting problems, the TWSO model is able to connect large gaps regardless of the geometry of the inpainting region and it would not introduce much blur to the inpainted image. As the proposed TWSO model is based on the variational framework, the ADMM can be adapted to solve the model efficiently. Extensive numerical results show that the new TWSO model outperforms the state-of-the-art approaches for both image inpainting and denoising. 

The contributions of the paper are twofold: 1) a new second order variational image reconstruction model is proposed. To the best of our knowledge, this is the first time the Frobenius norm of the product of the Hessian matrix and a tensor has been used as a regulariser for variational image denoising and inpainting; 2) A fast ADMM algorithm is developed for image reconstruction based on a forward-backward finite difference scheme.

The rest of paper is organised as follows: Section \ref{TWSO} introduces the proposed TWSO model and the anisotropic tensor $\textbf{T}$; Section \ref{Discretisation} presents the discretisation of the differential operators used for ADMM based on a forward-backfard finite difference scheme; Section \ref{ADMM} describes the ADMM for solving the variational model efficiently. Section \ref{experiments} gives details of the experiments using the proposed TWSO model and the state-of-the-art approaches for image inpainting and denoising. Section \ref{conclusion} concludes the paper.    

\section{The TWSO Model for Image Processing}
\label{TWSO}

\subsection{The TWSO model}
In \cite{lysaker2003noise,bergounioux2010second}, the authors considered the following SOTV model for image processing
\begin{equation} \label{eq:SOTV}
\mathop {\min }\limits_u \left\{ {\frac{\eta }{2}||u - f||_2^2 + ||{\nabla ^2}u|{|_1}} \right\},
\end{equation}
where $\eta>0$ is a regularisation parameter. ${\nabla ^2}u$ is the second order Hessian matrix of the form
\begin{equation} \label{eq:Hess}
{\nabla ^2}u = \left( \begin{array}{l}
\partial _x \partial _x u \;\; \partial _y \partial _x u\\
\partial _x \partial _y u \;\; \partial _y \partial _y u
\end{array} \right),
\end{equation}
and $||{\nabla ^2}u|{|_1}$ in (\ref{eq:SOTV}) is the Frobenius norm of the Hessian matrix (\ref{eq:Hess}). This high order model (\ref{eq:SOTV}) is able to remove the staircase artefact associated with the FOTV for image denoising, but it can blur object edges in the image. For inpainting, as investigated in \cite{papafitsoros2013combined,papafitsoros2014combined}, though the SOTV has the ability to connect large gaps in the image, such ability depends on the geometry of the inpainting region, and it can blur the inpainted image. In order to remedy these side effects in both image impaiting and denoising, we propose a more flexible and generalised variational model, i.e., the Tensor-Weighted Second Order (TWSO) model, that can take advantages of both the tensor and the second order derivative. Specifically, the TWSO model is defined as
\begin{equation} \label{eq:TWSO}
\mathop {\min }\limits_u \left\{ {\frac{\eta }{p}||{1 _\Gamma }\left( {u - f} \right)||_p^p + ||{\bf{T}}{\nabla ^2}u|{|_1}} \right\},
\end{equation}
where $p \in \left\{1,2\right\}$ denotes the $L^1$ and $L^2$ data fidelity terms (i.e. ${||{1 _\Gamma }\left( {u - f} \right)||_1}$ and ${||{1 _\Gamma }\left( {u - f} \right)||_2^2}$), and $\Gamma$ is a subset of ${\Omega}$ (i.e. $\Gamma  \subset \Omega \subset {{R}^2}$). For image processing applications, the $\Omega$ is normally a rectangle domain. For image inpainting, $f$ is the given image in $\Gamma=\Omega \backslash D$, where $D \subset \Omega$ is the inpainting region with missing or degraded information. The values of $f$ on the boundaries of $D$ need to be propagated into the inpainting region via minimisation of the weighted regularisation term of the TWSO model. For image denoising, $f$ is the noisy image and $\Omega=\Gamma$. In this case, the choice of $p$ depends on the type of noise found in $f$, e.g. $p=2$ for Gaussian noise while $p=1$ for impulsive noise. 

In the regularisation term ${||{\bf{T}}{\nabla ^2}u|{|_1}}$ of (\ref{eq:TWSO}), $\textbf{T}$ is a symmetric positive semi-definite 2$\times$2 diffusion tensor whose four components are $\textbf{T}_{11}$, $\textbf{T}_{12}$, $\textbf{T}_{21}$, and $\textbf{T}_{22}$. They are computed from the input image $f$. It is worth pointing out that the regularisation term ${||{\bf{T}}{\nabla ^2}u|{|_1}}$ is the Frobenius norm of a 2 by 2 tensor $\textbf{T}$ multiplied by a 2 by 2 Hessian matrix ${\nabla ^2}u$, which has the form of
\begin{equation} \label{eq:coupleHess}
\left( \begin{array}{l}
{\textbf{T}_{11}}{\partial _x}{\partial _x}u + {\textbf{T}_{12}}{\partial _x}{\partial _y}u\;\;{\textbf{T}_{11}}{\partial _y}{\partial _x}u + {\textbf{T}_{12}}{\partial _y}{\partial _y}u\\
{\textbf{T}_{21}}{\partial _x}{\partial _x}u + {\textbf{T}_{22}}{\partial _x}{\partial _y}u\;\;{\textbf{T}_{21}}{\partial _y}{\partial _x}u + {\textbf{T}_{22}}{\partial _y}{\partial _y}u
\end{array} \right),
\end{equation}
where the two orthogonal eigenvectors of $\textbf{T}$ span the rotated coordinate system in which the gradient of the input image is computed and as such $\textbf{T}$ can introduce orientations to the bounded Hessian regulariser ${||{{\nabla ^2}u}||_1}$. The eigenvalues of the tensor measure the degree of anisotropy in the regulariser and weight the four second order derivatives in ${{\nabla ^2}u}$ in the two directions given by the eigenvectors of the structure tensor introduced in \cite{weickert1998anisotropic}. As a result, the new tensor-weighted regulariser ${||{\bf{T}}{\nabla ^2}u|{|_1}}$ is powerful for image image inpainting and denoising, as illustrated in the experimental section. In the next section, we shall introduce the derivation of the tensor $\textbf{T}$. 

\subsection{Tensor Estimation}
In \cite{weickert1998anisotropic}, the author defined the structure tensor $J_\rho$ of an image $u$ 
\begin{equation} \label{eq:ST}
J_\rho\left( {\nabla u_{\sigma}} \right) = {K_\rho}*\left( {\nabla {u_\sigma } \otimes \nabla {u_\sigma }} \right), 
\end{equation}
where ${K_\rho}$ is a Gaussian kernel whose standard deviation is $\rho$ and $\nabla u_\sigma$ is the smoothed version of the gradient convolved by $K_\sigma$. The use of $\left({\nabla {u_\sigma } \otimes \nabla {u_\sigma }} \right):=\nabla {u_\sigma }\nabla u_\sigma ^{\rm{T}}$ as a structure descriptor is to make $J_\rho$ insensitive to noise but sensitive to change in orientation. The structure tensor $J_\rho$ is positive semi-definite and has two orthonormal eigenvectors $v_1\;||\;\nabla u_\sigma$ (in the direction of gradient) and $v_2\;||\;\nabla u_\sigma$ (in the direction of the isolevel lines). The corresponding eigenvalues $\mu_1$ and $\mu_2$ can be calculated from
\begin{equation}
{\mu_{1,2}} = \frac{1}{2}\left( {{j_{11}} + {j_{22}} \pm \sqrt {{{\left( {{j_{11}} - {j_{22}}} \right)}^2} + 4j_{12}^2} } \right),
\end{equation}
where $j_{11}$, $j_{12}$ and $j_{22}$ are the components of $J_\rho$. They are given as 
\begin{equation} \label{eq:Jp}
{j_{11}} = {K_\rho}*{\left( {{\partial _x}{u_\sigma }} \right)^2},\;{j_{12}} = {j_{21}} = {K_\rho}*\left( {{\partial _x}{u_\sigma }{\partial _y}{u_\sigma }} \right),\;{j_{22}} = {K_\rho}*{\left( {{\partial _y}{u_\sigma }} \right)^2}.
\end{equation}
The eigenvalues of $J_\rho$ describe the $\rho$-averaged contrast in the eigendirections, meaning: if $\mu_1=\mu_2=0$, the image is in homogeneous area; if $\mu_1\gg\mu_2=0$, it is on a straight line; and if $\mu_1>\mu_2\gg0$, it is at objects' corner. Based on the eigenvalues, we can define the following local structural coherence quantity
\begin{equation}
Coh=(\mu_1-\mu_2)^2=(j_{11}-j_{22})^2+4j_{12}^2.
\end{equation}
This quantity is large for linear structures and small for homogeneous areas in the image. With the derived structure tensor (\ref{eq:ST}) we define a new tensor ${{\bf{T}}} = D\left( {J_\rho\left( {\nabla {u_{\sigma}}} \right)} \right)$ whose eigenvectors are parallel to the ones of $J_\rho(\nabla u_{\sigma})$ and its eigenvalues $\lambda_1$ and $\lambda_2$ are chosen depending on different image processing applications. 
For denoising problems, we need to prohibit the diffusion across image edges and encourage strong diffusion along edges. We therefore consider the following two diffusion coefficients for the eigenvalues
\begin{equation} \label{eq:edgePresEigVal}
{\lambda _1} = \left\{ \begin{array}{*{20}{l}}
1 &\;s \le 0\;\\
1 - {e^{\frac{{ - 3.31488}}{{{{\left( {{s \mathord{\left/
 {\vphantom {s C}} \right.
 \kern-\nulldelimiterspace} C}} \right)}^8}}}}}&\;s > 0
\end{array} \right.,\;\;\;\;{\lambda _2} = 1,
\end{equation}
with $s:={\left| {\nabla {u_\sigma }} \right|}$ the gradient magnitude and $C$ the contrast parameter.
For inpainting problems, we want to preserve linear structures and thus regularisation along isophotes of the image is appropriate. We therefore consider the weights that Weickert \cite{weickert1998anisotropic} used for enhancing the coherence of linear structures. With $\mu_1$, $\mu_2$ being the eigenvalues of $J_\rho$ as before, we define
\begin{equation}\label{eq:struPresEigVal}
{{\lambda _1} = \gamma ,\;\;\;\;{\lambda _2} = \left\{ {\begin{array}{*{20}{l}}
\gamma &\;{{\mu _1} = {\mu _2}}\\
{\gamma  + \left( {1 - \gamma } \right){e^{ - \frac{C}{{Coh}}}}}&\;{{\rm{else}}}
\end{array}} \right.},
\end{equation} 
where $\gamma \in (0, 1)$, $\gamma \ll 1$. The constant $\gamma$ determines how steep the exponential function is. The structure threshold $C$ affects how the approach interprets local structures. The larger the parameter value is, the more coherent the method will be. With these eigenvalues, the regularisation is stronger in the neighbourhood of coherent structures (note that $\rho$ determines the radius of neighbourhood) and weaker in homogeneous areas, at corners, and in incoherent areas of the image.

\section{Discretisation of differential operators}
\label{Discretisation}
In order to implement ADMM for the proposed TWSO model, it is necessary to discretise the derivatives involved. Let $\Omega$ denote the two dimensional grey scale image space of size MN, and $x$ and $y$ denote the coordinates along image column and row direction respectively. The discrete second order derivatives of $u$ at point $\left( {i,j} \right)$ along $x$ and $y$ directions can be then written as
\begin{equation}
\partial _x^ + \partial _x^ - {u_{i,j}} = \partial _x^ - \partial _x^ + {u_{i,j}} = \left\{ \begin{array}{ll}
{u_{i,N}} - 2{u_{i,j}} + {u_{i,j + 1}} & \rm{if}\;1 \le i \le M,\;j = 1\\
{u_{i,j - 1}} - 2{u_{i,j}} + {u_{i,j + 1}} & \rm{if}\;1 \le i \le M,\;1 < j < N\;\\
{u_{i,j - 1}} - 2{u_{i,j}}\; + {u_{i,1}} & \rm{if}\;1 \le i \le M,\;j = N\;
\end{array} \right., \label{eq:FxBx(BxFx)}
\end{equation}
\begin{equation}
\partial _y^ + \partial _y^ - {u_{i,j}} = \partial _y^ - \partial _y^ + {u_{i,j}} = \left\{ \begin{array}{ll}
{u_{M,j}} - 2{u_{i,j}} + {u_{i + 1,j}} & \rm{if}\;i = 1,\;1 \le j \le N\;\\
{u_{i - 1,j}} - 2{u_{i,j}} + {u_{i + 1,j}} & \rm{if}\;1 < i < M,\;1 \le j \le N\;\\
{u_{i - 1,j}} - 2{u_{i,j}} + {u_{1,j}} & \rm{if}\;i = M,\;1 \le j \le N\;
\end{array} \right., \label{eq:FyBy(ByFy)}
\end{equation}
\begin{equation}
\partial _x^ + \partial _y^ + {u_{i,j}} = \partial _y^ + \partial _x^ + {u_{i,j}} = \left\{ \begin{array}{ll}
{u_{i,j}} - {u_{i + 1,j}} - {u_{i,j + 1}} + {u_{i + 1,j + 1}} & \rm{if}\;1 \le i < M,\;1 \le j < N\\
{u_{i,j}} - {u_{1,j}} - {u_{i,j + 1}} + {u_{1,j + 1}} & \rm{if}\;i = M,\;1 \le j < N\\
{u_{i,j}} - {u_{i + 1,j}} - {u_{i,1}} + {u_{i + 1,1}} & \rm{if}\;1 \le i < M,\;j = N\\
{u_{i,j}} - {u_{1,j}} - {u_{i,1}} + {u_{1,1}} & \rm{if}\;i = M,\;j = N\;
\end{array} \right.,\label{eq:FxFy(FyFx)}
\end{equation}
\begin{equation}
\partial _x^ - \partial _y^ - {u_{i,j}} = \partial _y^ - \partial _x^ - {u_{i,j}} = \left\{ \begin{array}{ll}
{u_{i,j}} - {u_{i,N}} - {u_{M,j}} + {u_{M,N}} & \rm{if}\;i = 1,\;j = 1\\
{u_{i,j}} - {u_{i,j - 1}} - {u_{M,j}} + {u_{M,j - 1}} & \rm{if}\;i = 1,\;1 < j \le N\\
{u_{i,j}} - {u_{i,N}} - {u_{i - 1,j}} + {u_{i - 1,N}} & \rm{if}\;1 < i \le M,\;j = 1\\
{u_{i,j}} - {u_{i,j - 1}} - {u_{i - 1,j}} + {u_{i - 1,j - 1}} & \rm{if}\;1 < i \le M,\;1 < j \le N
\end{array} \right.. \label{eq:BxBy(ByBx)}
\end{equation}

\begin{figure}[h!] 
\centering
\subfigure[\tiny {$\partial _x^ + \partial _x^ - =\partial _x^ - \partial _x^ + $}]
{\includegraphics[width=0.2\textwidth]{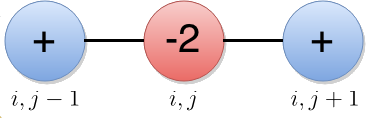}} \;\;\;
\subfigure[\tiny {$\partial _y^ + \partial _y^ -  = \partial _y^ - \partial _y^ +$}]
{\includegraphics[height=0.2\textwidth]{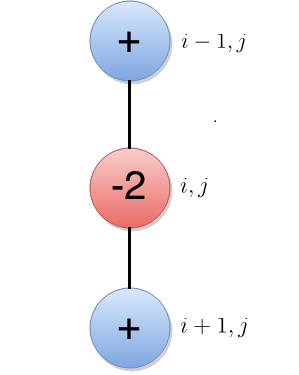}} \;\;\;
\subfigure[\tiny {$\partial _x^ + \partial _y^ +  = \partial _y^ + \partial _x^ + $}]
{\includegraphics[height=0.17\textwidth]{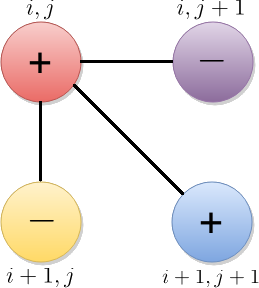}} \;\;\; 
\subfigure[\tiny {$\partial _x^ - \partial _y^ -  = \partial _y^ - \partial _x^ - $}]
{\includegraphics[height=0.17\textwidth]{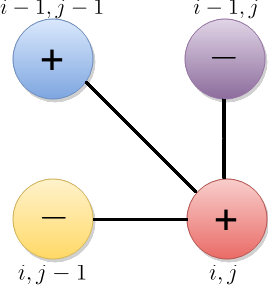}} \;\;\;
\caption{Discrete second order derivatives.}
\label{fig:discreteIll}
\end{figure}

\noindent Figure~\ref{fig:discreteIll} summarises these discrete differential operators. Based on the above forward-backward finite difference scheme, the second order Hessian matrix $\nabla^2 u$ in (\ref{eq:Hess}) can be descretised as
\begin{equation} \label{eq:disHess}
{\nabla ^2}u = \left( \begin{array}{l}
\partial _x^ - \partial _x^ + u \;\; \partial _y^ + \partial _x^ + u\\
\partial _x^ + \partial _y^ + u \;\; \partial _y^ - \partial _y^ + u
\end{array} \right).
\end{equation}
In (\ref{eq:FxBx(BxFx)})-(\ref{eq:disHess}), we assume that $u$ satisfies the periodic boundary condition so that the fast Fourier transform (FFT) solver can be applied to solve (\ref{eq:TWSO}) analytically. The numerical approximation of the second order divergence operator $div^2$ is based on the following expansion
\begin{equation}
di{v^2}{\left( \textbf{P}\right)} = \partial _x^ + \partial _x^ - {\left( {{\textbf{P}_1}} \right)} + \partial _y^ - \partial _x^ - {\left( {{\textbf{P}_2}} \right)} + \partial _x^ - \partial _y^ - {\left( {{\textbf{P}_3}} \right)} + \partial _y^ + \partial _y^ - {\left( {{\textbf{P}_4}} \right)}. \label{eq:SOdivergence}
\end{equation}
where $\textbf{P}$ is a 2$\times$2 matrix whose four components are ${\textbf{P}_1}$, ${\textbf{P}_2}$, ${\textbf{P}_3}$ and ${\textbf{P}_4}$, respectively. For more detailed description of the discretisation of other high order differential operators, we refer the reader to \cite{duan2016edge,lu2016implementation}.

Finally, we address the implementation problem of the first order derivatives of $u_\sigma$ in (\ref{eq:Jp}). Since the differentiation and convolution are commutative, we can take the derivative and smooth the image in either order. In this sense, we have 
\begin{equation}
{{\partial _x}{u_\sigma } = {\partial _x}{K_\sigma }*u,\;\;\;\;{\partial _y}{u_\sigma } = {\partial _y}{K_\sigma }*u}.
\end{equation}
Alternatively, the central finite difference scheme can be used to approximate ${\partial _x}{u_\sigma }$ and ${\partial _y}{u_\sigma }$ to satisfy the rotation-invariant property. Once all necessary discretisation is done, the numerical computation can be implemented. 

\section{Numerical Optimisation Algorithm}
\label{ADMM}
It is nontrivial to directly solve the TWSO model due to the facts: i) it is nonsmooth; ii) it couples the tensor $\textbf{T}$ and Hessian matrix ${\nabla ^2}u$ in ${||{\bf{T}}{\nabla ^2}u|{|_1}}$ as shown in (\ref{eq:coupleHess}), making the resulting high order Euler-Lagrange equation drastically difficult to discretise to solve computationally. To address these two difficulties, we present an efficient numerical algorithm based on ADMM to minimise the variational model (\ref{eq:TWSO}). 

\subsection{Alternating direction method of multipliers (ADMM)}
ADMM is combines the decomposability of the dual ascent with superior convergence properties of the method of multipliers. Recent research \cite{boyd2011distributed} unveils that ADMM is also closely related to Douglas-Rachford splitting, Spingarn's method of partial inverses, Dykstra's alternating projections, split Bregman iterative algorithm etc. Given a constrained optimisation problem
\begin{equation} \label{eq:ADMM}
\mathop {\min }\limits_{{\rm{x}},{\rm{z}}} \left\{ {f\left( {\rm{x}} \right) + g\left( {\rm{z}} \right)} \right\}\:\:{\rm{s}}.{\rm{t}}.\:\:{\rm A}{\rm{x + }}{\rm B}{\rm{z = }}c,
\end{equation}
where ${\rm{x}} \in {{\rm{R}}^{{d_1}}}$, ${\rm{z}} \in {{\rm{R}}^{{d_2}}}$, ${\rm A} \in {{\rm{R}}^{m \times {d_1}}}$, ${\rm B} \in {{\rm{R}}^{m \times {d_2}}}$, $c \in {{\rm{R}}^m}$, and both ${f\left(  \cdot  \right)}$ and ${g\left(  \cdot  \right)}$ are assumed to be convex. The augmented Lagrange function of problem (\ref{eq:ADMM}) can be written as
\begin{equation} \label{eq:ADMM1}
\mathcal{L_A}\left( {{\rm{x}},{\rm{z}};\rho } \right) = f\left( {\rm{x}} \right) + g\left( {\rm{z}} \right) + \frac{\beta }{2}||{\rm A}{\rm{x}} + {\rm B}{\rm{z}} - c - \rho ||_2^2,
\end{equation}
where $\rho$ is an augmented Lagrangian multiplier and $\beta>0$ is an augmented penalty parameter. At the $k$th iteration, ADMM attempts to solve problem (\ref{eq:ADMM1}) by iteratively minimising $\mathcal{L_A}$ with respect to $x$ and $z$, and updating $\rho$ accordingly. The resulting optimisation procedure is summarised in \hyperlink{alogrithm1}{\textbf{Algorithm 1}}.

\hypertarget{alogrithm1}{}
\begin{table}[h!] 
\centering
\begin{tabular}{p{9cm}}
\hline
\rowcolor{Gray}
\textbf{Algorithm 1}: ADMM \\
\hline
1: Initialization: Set $\beta>0$, $z^0$ and $\rho^0$.\\
2: \textbf{while} \textit{a stopping criterion is not satisfied} \textbf{do}\\
3: \quad ${{\rm{x}}^{k + 1}} = \mathop {\arg \min_{\rm{x}} } \left\{ {f\left( {\rm{x}} \right) + \frac{\beta }{2}||{\rm A}{\rm{x}} + {\rm B}{{\rm{z}}^k} - c - {\rho ^k}||_2^2} \right\}$.\\
4: \quad ${{\rm{z}}^{k + 1}} = \mathop {\arg \min_{\rm{z}} } \left\{ {g\left( {\rm{z}} \right) + \frac{\beta }{2}||{\rm A}{{\rm{x}}^{k + 1}} + {\rm B}{{\rm{z}}^k} - c - {\rho ^k}||_2^2} \right\}$.\\
5: \quad ${\rho ^{k + 1}} = {\rho ^k} - \left( {{\rm A}{{\rm{x}}^{k + 1}} + {\rm B}{{\rm{z}}^{k + 1}} - c} \right)$.\\
6: \textbf{end while}\\
\hline
\end{tabular}
\end{table} 

\subsection{Application of ADMM to solve the TWSO model}
We now use ADMM to solve the minimisation problem of the proposed TWSO model (\ref{eq:TWSO}). The basic idea of ADMM is to first split the original nonsmooth minimisation problem into several subproblems by introducing some auxiliary variables, and then solve each subproblem separately. This numerical algorithm benefits from both solution stability and fast convergence.

In order to implement ADMM, one scalar auxiliary variable $\tilde u$ and two 2$\times$2 matrix-valued auxiliary variables $\textbf{W}$ and $\textbf{V}$ are introduced to reformulate (\ref{eq:TWSO}) into the following constraint optimisation problem 
\begin{equation} \label{eq:ConstraintTWSO}
\mathop {\min }\limits_{\tilde u, u,{\bf{W}},{\bf{V}}} \left\{ {\frac{\eta }{p}||{1 _\Gamma }\left( {\tilde u - f} \right)||_p^p + ||{\bf{W}}|{|_1}} \right\}
\text{s.t.}\;\tilde u = u, \textbf{V} = {\nabla ^2}u,\textbf{W} = \textbf{T}\textbf{V},
\end{equation}
{where} $\textbf{W} = ( \begin{array}{l}
{\textbf{W}_{11}}\;{\textbf{W}_{12}}\\
{\textbf{W}_{21}}\;{\textbf{W}_{22}}
\end{array} )$, $\textbf{V} = ( \begin{array}{l}
{\textbf{V}_{11}}\;{\textbf{V}_{12}}\\
{\textbf{V}_{21}}\;{\textbf{V}_{22}}
\end{array} )$, $\textbf{d} = ( \begin{array}{l}
{\textbf{d}_{11}}\;{\textbf{d}_{12}}\\
{\textbf{d}_{21}}\;{\textbf{d}_{22}}
\end{array} )$ and $\textbf{b} = ( \begin{array}{l}
{\textbf{b}_{11}}\;{\textbf{b}_{12}}\\
{\textbf{b}_{21}}\;{\textbf{b}_{22}}
\end{array} )$. The constraints $\tilde u=u$ and $\textbf{W}=\textbf{TV}$ are applied to handle the non-smoothness of the $L^1$-norm of the data fidelity term ($p=1$) and the weighted regularisation term respectively, whilst $\textbf{V}=\nabla^2 u$ decouples $\textbf{T}$ and $\nabla^2 u$ in the TWSO. To guarantee an optimal solution, the above constrained problem (\ref{eq:ConstraintTWSO}) can be solved through ADMM summarised in \hyperlink{alogrithm1}{\textbf{Algorithm 1}}. Let $\mathcal{L_A}\left( {\tilde u, u,{\bf{W}},{\bf{V}};s,{\bf{d}},{\bf{b}}} \right)$ be the augmented Lagrange functional of (\ref{eq:ConstraintTWSO}), which is defined as follows
\begin{equation} \label{eq:SBM}
\begin{split}
\mathcal{L_A}\left( {\tilde u, u,{\bf{W}},{\bf{V}};s,{\bf{d}},{\bf{b}}} \right) = \frac{\eta }{p}||{1 _\Gamma }\left( {\tilde u - f} \right)||_p^p + ||{\bf{W}}||_1 + \frac{{{\theta _1}}}{2}||\tilde u - u - s||_2^2 \\+\frac{{{\theta _2}}}{2}||{\bf{V}} - {\nabla ^2}u - {\bf{d}}||_2^2 + \frac{{{\theta _3}}}{2}||{\bf{W}} - {\bf{TV}} - {\bf{b}}||_2^2,
\end{split}
\end{equation}
where $s$, $\textbf{d}$ and $\textbf{b}$ are the augmented Lagrangian multipliers, and $\theta_1$, $\theta_2$ and $\theta_3$ are positive penalty constants controlling the weights of the penalty terms. 

We will now decompose the optimisation problem (\ref{eq:SBM}) into four subproblems with respect to $\tilde u$, $u$, $\textbf{W}$ and $\textbf{V}$, and then update the Lagrangian multipliers $s$, $\textbf{d}$ and $\textbf{b}$ until the optimal solution is found and the process converges. 

1) $\tilde u-$\textit{subproblem}: This problem $\tilde u^{k+1} \leftarrow{\min }_{\tilde u}\mathcal{L_A}\left( {\tilde u, u^k,{\bf{W}}^k,{\bf{V}}^k;s^k,{\bf{d}}^k,{\bf{b}}^k} \right)$ can be solved by considering the following minimisation problem
\begin{equation} \label{eq:uba}
{{\tilde u}^{k + 1}} = \arg {\min _{\tilde u}}\left\{ {\frac{\eta }{p}||{1_\Gamma }\left( {\tilde u - f} \right)||_p^p + \frac{{{\theta _1}}}{2}||\tilde u - {u^k} - {s^k}||_2^2} \right\}.
\end{equation}
The solution of (\ref{eq:uba}) depends on the choice of $p$. Given the domain $\Gamma$ for image denoising or inpainting, the closed-form formulae for the minimisers $\tilde u^{k+1}$ under different conditions are  
\begin{equation} \label{eq:uba0}
 \left\{
\begin{array}{ll}
      {{\tilde u}^{k + 1}} = {{\left( {{1_\Gamma }\eta f + {\theta _1}\left( {{u^k} + {s^k}} \right)} \right)} \mathord{\left/
 {\vphantom {{\left( {{1_\Gamma }\eta f + {\theta _1}\left( {{u^k} + {s^k}} \right)} \right)} {\left( {{1_\Gamma }\eta  + {\theta _1}} \right)}}} \right.
 \kern-\nulldelimiterspace} {\left( {{1_\Gamma }\eta  + {\theta _1}} \right)}}   &\;\rm{if}\; \textit{p} = 2 \\
      {{\tilde u}^{k + 1}} = f + \max \left( {|{\psi ^k}| - \frac{{{1_\Gamma }\eta }}{{{\theta _1}}},0} \right) \circ sign\left( {{\psi ^k}} \right) & \;\rm{if}\; \textit{p}=1 \\
\end{array} 
\right.,
\end{equation}
where ${\psi ^k} = {u^k} + {s^k} - f$. $\circ$ and $sign$ symbols denote the component-wise multiplication and signum function, respectively.

2) $u-$\textit{subproblem}: We then solve $u$-subproblem ${u^{k + 1}} \leftarrow {\min _u}{{\cal L}_{\cal A}}\left( {\tilde u^{k+1},{u},{{\bf{W}}^k},} \right.$ $\left. {{{\bf{V}}^k};{s^k},{{\bf{d}}^k},{{\bf{b}}^k}} \right)$ by minimising the following problem.
\begin{equation}\label{eq:FFTFunctional}
{u^{k + 1}} = \arg {\min _u}\left\{ {\frac{{{\theta _1}}}{2}||{{\tilde u}^{k + 1}} - u - {s^k}||_2^2 + \frac{{{\theta _2}}}{2}||{{\bf{V}}^k} - {\nabla ^2}u - {{\bf{d}}^k}||_2^2} \right\},
\end{equation} 
whose closed-form can be obtained using the following FFT under the assumption of the circulant boundary condition (Note that to benefit from the fast FFT solver for image inpainting problems, the introduction of $\tilde u$ is compulsory due to the fact that ${\cal{F}}({1_{\Gamma}u})\ne {1_{\Gamma}{\cal{F}}(u})$)
\begin{equation} \label{eq:FFT}
{u^{k + 1}} = {{\cal{F}}^{ - 1}}\left( {\frac{{{\cal{F}}\left( {{\theta _1}\left( {{{\tilde u}^{k + 1}} - {s^k}} \right) + {\theta _2}di{v^2}\left( {{{\bf{V}}^k} - {{\bf{d}}^k}} \right)} \right)}}{{{\theta _1} + {\theta _2}{\cal{F}}\left( {di{v^2}{\nabla ^2}} \right)}}} \right),
\end{equation}
where $\cal{F}$ and ${\cal F}^{-1}$ respectively denote the discrete Fourier transform and inverse Fourier transform; $div^2$ is a second order divergence operator whose discrete form is defined in (\ref{eq:SOdivergence}); \enquote{---} stands for the pointwise division of matrices. The values of the coefficient matrix ${\cal{F}}( {di{v^2}{\nabla ^2}} )$ equal $4\left({\cos \frac{{2\pi q}}{N} + \cos \frac{{2\pi r}}{M} - 2}\right)$, where $M$ and $N$ respectively stand for the image width and height, and $r\in[0,M)$ and $q\in[0,N)$ are the frequencies in the frequency domain. Note that in addition to FFT, AOS and Gauss-Seidel iteration can be applied to minimise the problem (\ref{eq:FFTFunctional}) with very low cost.

3) $\textbf{W}-$\textit{subproblem}: We now solve the \textbf{W}-subproblem ${\textbf{W}^{k + 1}} \leftarrow {\min _\textbf{W}}{{\cal L}_{\cal A}}\left( {\tilde u^{k+1},} \right.$ $\left. {{u}^{k+1},{{\bf{W}}},{{\bf{V}}^k};{s^k},{{\bf{d}}^k},{{\bf{b}}^k}} \right)$. Note that the unknown matrix-valued variable \textbf{W} is componentwise separable, which can be effectively solved through the analytical shrinkage operation, also known as the soft generalised thresholding equation
\begin{equation} \nonumber
{{\bf{W}}^{k + 1}} = \arg {\min _{\bf{W}}}\left\{ {||{\bf{W}}|{|_1} + \frac{{{\theta _3}}}{2}||{\bf{W}} - {\bf{T}}{{\bf{V}}^k} - {{\bf{b}}^k}||_2^2} \right\},
\end{equation}
whose solution $\textbf{W}^{k+1}$ is given by 
\begin{equation} \label{eq:softThresholdingEquation}
{{\bf{W}}^{k + 1}}{\rm{ }} = \max \left( {\left| {{\bf{T}}{{\bf{V}}^k} + {{\bf{b}}^k}} \right| - \frac{1}{{{\theta _3}}},0} \right)\circ\frac{{{\bf{T}}{{\bf{V}}^k} + {{\bf{b}}^k}}}{{\left| {{\bf{T}}{{\bf{V}}^k} + {{\bf{b}}^k}} \right|}},
\end{equation}
with the convention that $0 \cdot \left( {{0 \mathord{\left/{\vphantom {0 0}} \right.\kern-\nulldelimiterspace} 0}} \right) = 0$.

4) The $\textbf{V}-$\textit{subproblem}: Given fixed $u^{k+1}$, $\textbf{W}^{k+1}$, $\textbf{d}^k$, $\textbf{b}^k$, the solution $\textbf{V}^{k+1}$ of the $\textbf{V}$-subproblem ${\textbf{V}^{k + 1}} \leftarrow {\min _\textbf{V}}{{\cal L}_{\cal A}}\left( {\tilde u^{k+1},} \right.$ $\left. {{u}^{k+1},{{\bf{W}}}^{k+1},{{\bf{V}}};{s^k},{{\bf{d}}^k},{{\bf{b}}^k}} \right)$ is equivalent to solving the following least-square optimisation problem
\begin{equation} \nonumber
{{\bf{V}}^{k + 1}} = \arg {\min _{\bf{V}}}\left\{ {\frac{{{\theta _2}}}{2}||{\bf{V}} - {\nabla ^2}{u^{k + 1}} - {{\bf{d}}^k}||_2^2 + \frac{{{\theta _3}}}{2}||{{\bf{W}}^{k + 1}} - {\bf{TV}} - {{\bf{b}}^k}||_2^2} \right\},
\end{equation}
which results in the following linear system with respect to each component in the variable $\textbf{V}^{k+1}$ 
\begin{equation} \label{eq:linearSystem}
\left\{ \begin{array}{l}
{\textbf{R}_{11}}\textbf{V}_{11}^{k + 1} + {\textbf{R}_{21}}\textbf{V}_{21}^{k + 1} = {\theta _2}( {\partial _x^{-}\partial _x^{+}{u^{k + 1}} + \textbf{d}_{11}^{k }} ) - \theta_3 ( {{\textbf{T}_{11}}{\textbf{Q}_{11}} + {\textbf{T}_{21}}{\textbf{Q}_{21}}} )\\
{\textbf{R}_{12}}\textbf{V}_{11}^{k + 1} + {\textbf{R}_{22}}\textbf{V}_{21}^{k + 1} = {\theta _2}( {\partial _x^{+}\partial _y^{+}{u^{k + 1}} + \textbf{d}_{21}^{k }} ) - \theta_3 ( {{\textbf{T}_{12}}{\textbf{Q}_{11}} + {\textbf{T}_{22}}{\textbf{Q}_{21}}} )\\
{\textbf{R}_{11}}\textbf{V}_{12}^{k + 1} + {\textbf{R}_{21}}\textbf{V}_{22}^{k + 1} = {\theta _2}( {\partial _y^{+}\partial _x^{+}{u^{k + 1}} + \textbf{d}_{12}^{k }} ) - \theta_3 ( {{\textbf{T}_{11}}{\textbf{Q}_{12}} + {\textbf{T}_{21}}{\textbf{Q}_{22}}} )\\
{\textbf{R}_{12}}\textbf{V}_{12}^{k + 1} + {\textbf{R}_{22}}\textbf{V}_{22}^{k + 1} = {\theta _2}( {\partial _y^{-}\partial _y^{+}{u^{k + 1}} + \textbf{d}_{22}^{k }} ) - \theta_3 ( {{\textbf{T}_{12}}{\textbf{Q}_{12}} + {\textbf{T}_{22}}{\textbf{Q}_{22}}} )
\end{array} \right.,
\end{equation}
where we define $\textbf{R}_{11}={\theta_3 (\textbf{T}_{11}^2 +\textbf{T}_{21}^2) + {\theta _2}}$, $\textbf{R}_{12}=\textbf{R}_{21}={\theta _3}\left( {{\textbf{T}_{11}}{\textbf{T}_{12}} + {\textbf{T}_{21}}{\textbf{T}_{22}}} \right)$, $\textbf{R}_{22}={\theta _3}\left( {\textbf{T}_{12}^2 + \textbf{T}_{22}^2} \right) + {\theta _2}$; ${{\bf{Q}}_{11}} = {\bf{b}}_{11}^k - {\bf{W}}_{11}^{k + 1}$, ${{\bf{Q}}_{12}} = {\bf{b}}_{12}^k - {\bf{W}}_{12}^{k + 1}$, ${{\bf{Q}}_{21}} = {\bf{b}}_{21}^k - {\bf{W}}_{21}^{k + 1}$ and ${{\bf{Q}}_{22}} = {\bf{b}}_{22}^k - {\bf{W}}_{22}^{k + 1}$. Note that the closed-forms of $\textbf{V}_{11}^{k+1}$ and $\textbf{V}_{21}^{k+1}$ can be calculated from the first two equations in (\ref{eq:linearSystem}), whilst the last two equations lead to the analytical solutions of $\textbf{V}_{12}^{k+1}$ and $\textbf{V}_{22}^{k+1}$. 

5) $s$, $\textbf{d}$ \textit{and} $\textbf{b}$ \textit{update}: At each iteration, we update the augmented Lagrangian multiplies $s$, $\textbf{d}$ and $\textbf{b}$ as shown from Step 09 to 11 in \hyperlink{alogrithm2}{\textbf{Algorithm 2}}.

In summary, an ADMM-based iterative algorithm was developed to decompose the original nonsmooth minimisation problem into four simple subproblems, each of which has a closed-form solution or can be efficiently solved using the existing numerical methods (i.e. FFT and shrinkage). The overall numerical optimisation algorithm of our proposed method for image restoration can be summarised in \hyperlink{alogrithm2}.

\hypertarget{alogrithm2}{}
\begin{table}[h!] 
\centering
\begin{tabular}{p{12cm}}
\hline
\rowcolor{Gray}
\textbf{Algorithm 2}: ADMM for the proposed TWSO model for image processing\\
\hline
\vspace{-5pt}
01: Input: $f$, $p$, $1_\Gamma$, $\eta$ and $(\theta_1,\theta_2,\theta_3)$.  \\
02: Initialise: $u=f$, ${\textbf{W}^0} = 0,\;{\textbf{V}^0} = 0,\;\textbf{d}^0 = 0,\;\textbf{b}^0 = 0$.\\
03: Calculate tensor $\textbf{T}$ using eigenvalues defined in (\ref{eq:edgePresEigVal}) or (\ref{eq:struPresEigVal}).\\
04: \textbf{while} \textit{some stopping criterion is not satisfied} \textbf{do}\\
05: \quad Compute $\tilde u^{k+1}$ according to (\ref{eq:uba0}).\\
06: \quad Compute $u^{k+1}$ according to (\ref{eq:FFT}).\\
07: \quad Compute ${\textbf{W}^{k+1}}$ according to (\ref{eq:softThresholdingEquation}).\\
08: \quad Compute ${\textbf{V}^{k+1}}$ according to (\ref{eq:linearSystem}).\\
09: \quad Update Lagrangian multiplier $s^{k+1}=s^{k}+u^{k+1}-\tilde u^{k+1}$. \\
10: \quad Update Lagrangian multiplier $\textbf{d}^{k+1}=\textbf{d}^{k}+\nabla^2 u^{k+1}-\textbf{V}^{k+1}$. \\
11: \quad Update Lagrangian multiplier $\textbf{b}^{k+1}=\textbf{b}^{k}+\textbf{TV}^{k+1}-\textbf{W}^{k+1}$.\\
12: {\textbf{end while}}\\
\hline
\end{tabular} 
\end{table}

\section{Numerical Experiments}
\label{experiments}
We conduct numerical experiments to compare the proposed model with state-of-the-art approaches for image impainting and denoising. The metrics for quantitative evaluation of the different methods are the peak signal-to-noise ratio (PSNR) and structure similarity index map (SSIM). The higher PSNR and SSIM a method obtains the better the method will be. In order to maximise the performance of all the compared methods, we carefully adjust their built-in parameters such that the resulting PSNR and SSIM by these methods are maximised. 

\subsection{Image Inpainting}
In this section, we test the capability of the proposed TWSO model for image inpainting. Several state-of-the-art inpainting methods will be compared with the TWSO model. They are the TV \cite{shen2002mathematical}, TGV \cite{schonlieb2015partial}, SOTV \cite{papafitsoros2013combined,papafitsoros2014combined}, and Euler's elastica \cite{shen2003euler,tai2011fast} inpainting models. In order to obtain a better inpainting result, we iteratively refine the diffusion tensor $\textbf{T}$ using the recovered image (i.e. $u^{k+1}$ in \hyperlink{alogrithm2}{\textbf{Algorithm 2}}). This refinement is crucial as it will provide more accurate tensor information for the next round iteration, thus leading to more pleasant inpainting results. We denote the inpainting domain as $D$ in the following experiments. The region $\Gamma$ in equation (\ref{eq:TWSO}) is $\Omega \backslash D$. We use $p=2$ for all examples in image inpainting.

\begin{figure}[h!] 
\centering  
\subfigure{\includegraphics[width=0.12\textwidth]{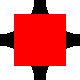}}
\subfigure{\includegraphics[width=0.12\textwidth]{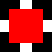}}
\subfigure{\includegraphics[width=0.12\textwidth]{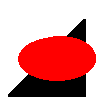}}
\subfigure{\includegraphics[width=0.12\textwidth]{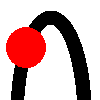}}
\subfigure{\includegraphics[width=0.12\textwidth]{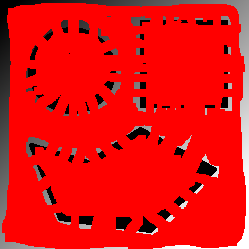}}
\subfigure{\includegraphics[width=0.12\textwidth]{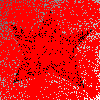}}\\
\vspace{-10pt}
\subfigure{\includegraphics[width=0.12\textwidth]{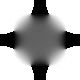}}
\subfigure{\includegraphics[width=0.12\textwidth]{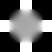}}
\subfigure{\includegraphics[width=0.12\textwidth]{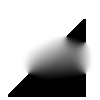}}
\subfigure{\includegraphics[width=0.12\textwidth]{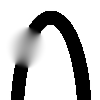}}
\subfigure{\includegraphics[width=0.12\textwidth]{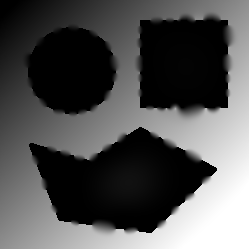}}
\subfigure{\includegraphics[width=0.12\textwidth]{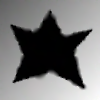}}\\
\vspace{-10pt}
\subfigure{\includegraphics[width=0.12\textwidth]{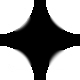}}
\subfigure{\includegraphics[width=0.12\textwidth]{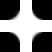}}
\subfigure{\includegraphics[width=0.12\textwidth]{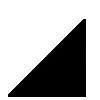}}
\subfigure{\includegraphics[width=0.12\textwidth]{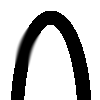}}
\subfigure{\includegraphics[width=0.12\textwidth]{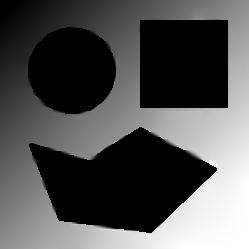}}
\subfigure{\includegraphics[width=0.12\textwidth]{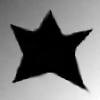}}\\
\vspace{-10pt}
\caption{Performance comparison of the SOTV and TWSO models on some degraded images. The red regions in the first row are inpainting domains. The second and third rows show the corresponding inpainting results by the SOTV and the proposed TWSO models, respectively.}
\label{fig:inpainting0}
\end{figure}
Figure~\ref{fig:inpainting0} illustrates the importance of the tensor $\textbf{T}$ in the proposed TWSO model. We compare the inpainted results by SOTV and TWSO models on some degraded images shown in the first row. From the second and third rows, it is clear that the TWSO performs much better than SOTV. SOTV introduces blurring to the inpainted regions, while TWSO recovers these shapes very well without causing much blurring. In addition to the capability of impainting large gaps, the TWSO can make smooth interpolation along the level curves of images on the inpainting domain. This validates the effectiveness of the tensor in TWSO for inpainting.

\begin{figure}[h!] 
\vspace{-10pt}
\centering  
{\includegraphics[width=0.12\textwidth]{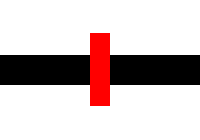}}
{\includegraphics[width=0.12\textwidth]{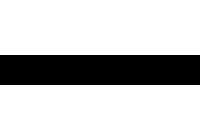}}
{\includegraphics[width=0.12\textwidth]{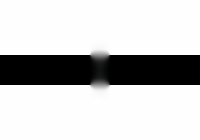}}
{\includegraphics[width=0.12\textwidth]{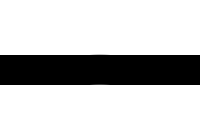}}
{\includegraphics[width=0.12\textwidth]{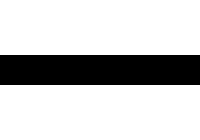}}
{\includegraphics[width=0.12\textwidth]{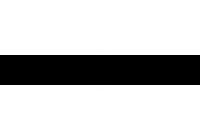}}\\
\vspace{-10pt}
{\includegraphics[width=0.12\textwidth]{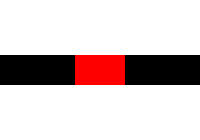}}
{\includegraphics[width=0.12\textwidth]{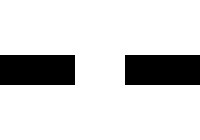}}
{\includegraphics[width=0.12\textwidth]{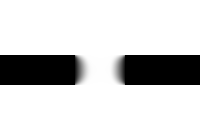}}
{\includegraphics[width=0.12\textwidth]{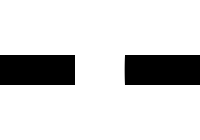}}
{\includegraphics[width=0.12\textwidth]{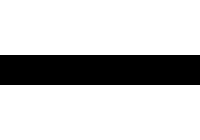}}
{\includegraphics[width=0.12\textwidth]{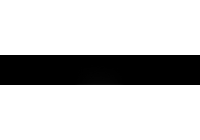}}\\
\vspace{-10pt}
{\includegraphics[width=0.12\textwidth]{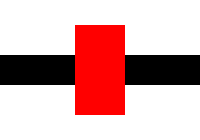}}
{\includegraphics[width=0.12\textwidth]{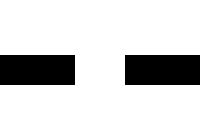}}
{\includegraphics[width=0.12\textwidth]{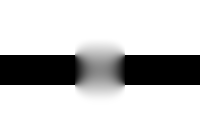}}
{\includegraphics[width=0.12\textwidth]{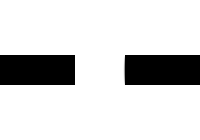}}
{\includegraphics[width=0.12\textwidth]{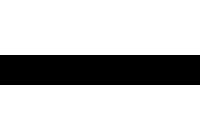}}
{\includegraphics[width=0.12\textwidth]{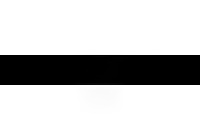}}\\
\vspace{-10pt}
\subfigure[]{\includegraphics[width=0.12\textwidth]{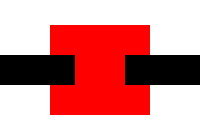}}
\subfigure[]{\includegraphics[width=0.12\textwidth]{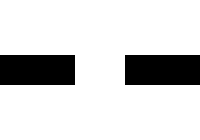}}
\subfigure[]{\includegraphics[width=0.12\textwidth]{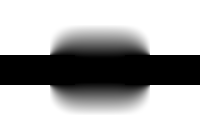}}
\subfigure[]{\includegraphics[width=0.12\textwidth]{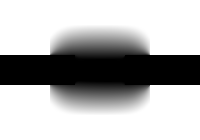}}
\subfigure[]{\includegraphics[width=0.12\textwidth]{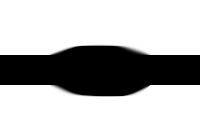}}
\subfigure[]{\includegraphics[width=0.12\textwidth]{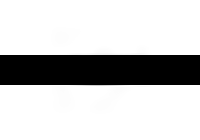}}\\
\vspace{-10pt}
\caption{Testing on a black stripe with different geometry of the inpainting domain. (a): Images with red inpainting regions. From top to bottom, the last three inpainting gaps have same width but with different geometry; (b): TV results; (c): SOTV results; (d): TGV results; (e): Euler's elastica results; (f): TWSO results.}
\label{fig:inpainting1}
\end{figure}

In Figure~\ref{fig:inpainting1}, we show how different geometry of the inpainting region affects the final results by the different methods. Column (b) illustrates that TV performs satisfactorily if the inpainting area is thin. However, it fails in images that contain large gaps. Column (c) indicates that the inpainting result by SOTV depends on the geometry of the inpainting region. It is able to impaint large gaps but at the price of introducing blurring. Mathematically, TGV is a combination of TV and SOTV, so its inpainted results, as shown in Column (d), are similar to those by TV and SOTV. TGV results also seem to depend on the geometry of the inpainting area. The last column shows that TWSO interpolates the images perfectly without any blurring and regardless of the inpainting region geometry. TWSO is even slightly better than Euler's elastica, which has proven to be very effective in dealing with larger gaps.

\begin{figure}[h!] 
\vspace{-5pt}
\centering  
\subfigure{\includegraphics[width=0.13\textwidth]{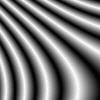}}
\subfigure{\includegraphics[width=0.13\textwidth]{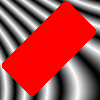}}
\subfigure{\includegraphics[width=0.13\textwidth]{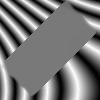}}
\subfigure{\includegraphics[width=0.13\textwidth]{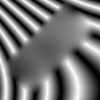}}
\subfigure{\includegraphics[width=0.13\textwidth]{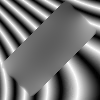}}
\subfigure{\includegraphics[width=0.13\textwidth]{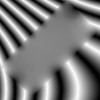}}
\subfigure{\includegraphics[width=0.13\textwidth]{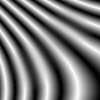}}\\
\vspace{-10pt}
\subfigure{\includegraphics[width=0.13\textwidth]{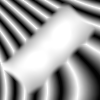}}
\subfigure{\includegraphics[width=0.13\textwidth]{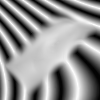}}
\subfigure{\includegraphics[width=0.13\textwidth]{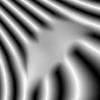}}
\subfigure{\includegraphics[width=0.13\textwidth]{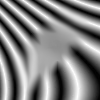}}
\subfigure{\includegraphics[width=0.13\textwidth]{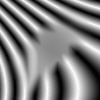}}
\subfigure{\includegraphics[width=0.13\textwidth]{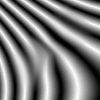}}
\subfigure{\includegraphics[width=0.13\textwidth]{Figure/smoothInpainting/TWSO}}\\
\vspace{-10pt}
\caption{Inpainting a smooth image with large gaps. First row from left to right are original image, degraded image, TV result, SOTV result, TGV result, Euler's elastica result, and TWSO result, respectively; Second row shows the intermediate results obtained by the TWSO model.}
\label{fig:inpainting2}
\vspace{-15pt}
\end{figure}

We now show another inpainting example of a smooth image with large gaps to further illustrate the effectiveness of the proposed TWSO model. As seen from Figure~\ref{fig:inpainting2}, all the methods except the TWSO model fail in correctly impainting the large gaps in the image. Note that TWSO integrates the tensor $\textbf{T}$ with its eigenvalues defined in (\ref{eq:struPresEigVal}), which makes it suitable for restoring linear structures, as shown in the second row of Figure~\ref{fig:inpainting2}. 

\begin{figure}[h!] 
\centering  
\subfigure{\includegraphics[width=0.15\textwidth]{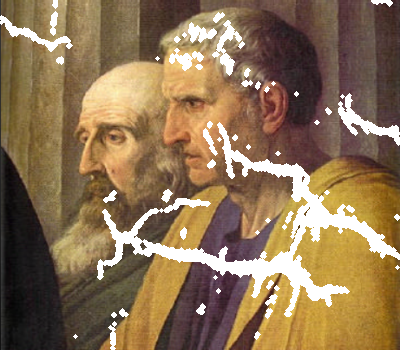}}
\subfigure{\includegraphics[width=0.15\textwidth]{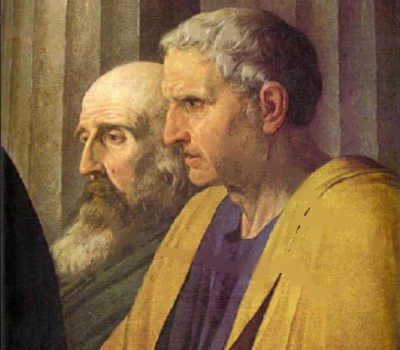}}
\subfigure{\includegraphics[width=0.15\textwidth]{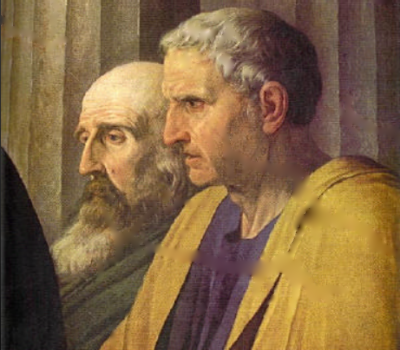}}
\subfigure{\includegraphics[width=0.15\textwidth]{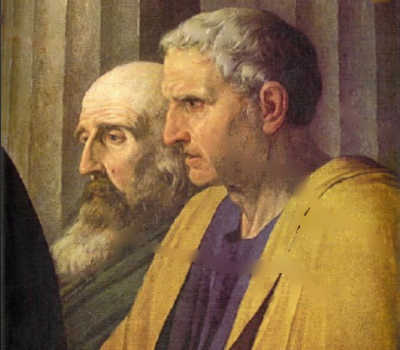}}
\subfigure{\includegraphics[width=0.15\textwidth]{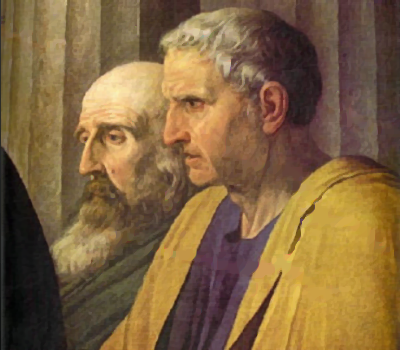}}
\subfigure{\includegraphics[width=0.15\textwidth]{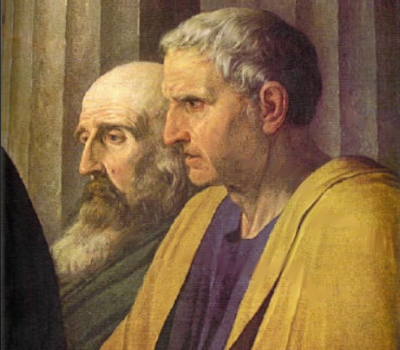}}\\
\vspace{-10pt}
\subfigure{\includegraphics[width=0.15\textwidth]{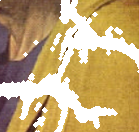}}
\subfigure{\includegraphics[width=0.15\textwidth]{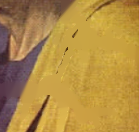}}
\subfigure{\includegraphics[width=0.15\textwidth]{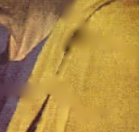}}
\subfigure{\includegraphics[width=0.15\textwidth]{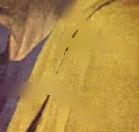}}
\subfigure{\includegraphics[width=0.15\textwidth]{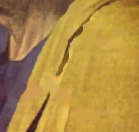}}
\subfigure{\includegraphics[width=0.15\textwidth]{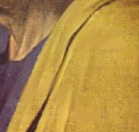}}\\
\vspace{-10pt}
\caption{Real example inpainting. ``Cornelia, mother of the Gracchi" by J. Suvee (Louvre). From left to right are degraded image for inpainting, TV result, SOTV result, TGV result, Euler's elastica result and TWSO result; The second row shows the local magnification of the corresponding results in the first row.}
\label{fig:inpainting3}
\end{figure}

Figure~\ref{fig:inpainting3} is a comparison of all the methods inpainting a real image. Though TV inpainting produces piecewise constant reconstruction in the inpainting domain, its result seems to be more satisfactory than those of the SOTV and TGV. As seen from the second row, neither TV nor TGV is able to connect the gaps on the cloth. SOTV interpolates some gaps but blurs the connection region. Euler's elastica performs better than TV, SOTV and TGV, but no better than the proposed TWSO model. Table~\ref{tb:table3} shows that the TWSO has is the most accurate among all the methods compared for the examples in Figure~\ref{fig:inpainting1}-\ref{fig:inpainting3}. 

\begin{table}[h!]
\caption{Comparison of PSNR and SSIM of different methods on Figure~\ref{fig:inpainting1}-\ref{fig:inpainting3}.}
\centering
\vspace{-10pt}
\resizebox{\columnwidth}{!}{
\begin{tabular}{|c|c|c|c|c|c|c|c|c|c|c|c|c|}
\hline
                           & \multicolumn{6}{c|}{PSNR test}                                                                                & \multicolumn{6}{c|}{SSIM value}                                                                               \\ \hline
\multirow{2}{*}{Figure \#} & \multicolumn{4}{c|}{Figure~\ref{fig:inpainting1}}                         & \multirow{2}{*}{Figure ~\ref{fig:inpainting2}} & \multirow{2}{*}{Figure~\ref{fig:inpainting3}} & \multicolumn{4}{c|}{Figure~\ref{fig:inpainting1}}                         & \multirow{2}{*}{Figure~\ref{fig:inpainting2}} & \multirow{2}{*}{Figure~\ref{fig:inpainting3}} \\ \cline{2-5} \cline{8-11}
                           & 1st row & 2nd row & 3rd row & 4th row &                           &                            & 1st row & 2nd row & 3rd row & 4th row &                           &                           \\ \hline
Degraded                   & 18.5316     & 12.1754     & 11.9285     & 11.6201     & 8.1022                    & 14.9359                   & 0.9418      & 0.9220  & 0.8778      & 0.8085      & 0.5541                    & 0.8589                    \\ \hline
TV                         & \textbf{Inf}         & 12.7107     & 12.7107     & 12.7107     & 14.4360                   & 35.7886                   & \textbf{1.0000}      & 0.9230      & 0.9230      & 0.9230      & 0.6028                    & 0.9936                    \\ \hline
SOTV                       & 28.7742     & 14.1931     & 18.5122     & 12.8831     & 14.9846                   & 35.6840                   & 0.9739      & 0.9289      & 0.9165      & 0.8118      & 0.6911                    & 0.9951                    \\ \hline
TGV                        & 43.9326     & 12.7446     & 12.7551     & 13.4604     & 14.0600                   & 36.0854                   & 0.9995      & 0.9232      & 0.9233      & 0.8167      & 0.6014                    & 0.9941                    \\ \hline
Euler's elastica           & 70.2979     & 54.0630     & 51.0182     & 13.0504     & 14.8741                   & 34.3064                   & \textbf{1.0000}      & 0.9946      & 0.9982      & 0.8871      & 0.6594                    & 0.9878                    \\ \hline
TWSO                       & \textbf{Inf}         & \textbf{Inf}         & \textbf{Inf}         & \textbf{43.3365}     & \textbf{27.7241}                   & \textbf{41.1400}                  & \textbf{1.0000}      & \textbf{1.0000}      & \textbf{1.0000}      & \textbf{0.9992}      & \textbf{0.9689}                    & \textbf{0.9976}                    \\ \hline
\end{tabular}}
\label{tb:table3}
\end{table}
 
\begin{table}[h!]
\caption{Mean and standard deviation (SD) of PSNR and SSIM calculated using 5 different methods for image inpainting over 100 images from the Berkeley database BSDS500 with 4 different random masks.}
\vspace{-10pt}
\resizebox{\columnwidth}{!}{
\begin{tabular}{|c||c|c|c|c||c|c|c|c|}
\hline
                                                        & \multicolumn{4}{c||}{PSNR value (Mean$\pm$ SD)}   & \multicolumn{4}{c|}{SSIM value (Mean$\pm$ SD)} \\ \hline \hline                         
Missing      & 40\%         & 60\%         & 80\%         & 90\%         & 40\%        & 60\%        & 80\%        & 90\%      \\ \hline
Degraded                                                & 9.020$\pm$0.4208 & 7.250$\pm$0.4200 & 6.010$\pm$0.4248 & 5.490$\pm$0.4245 & 0.05$\pm$0.0302 & 0.03$\pm$0.0158 & 0.01$\pm$0.0065 & 0.007$\pm$0.003 \\ \hline
TV                                                      & 31.99$\pm$3.5473 & 28.68$\pm$3.3217 & 24.61$\pm$2.7248 & 20.94$\pm$2.2431 & 0.92$\pm$0.0251 & 0.84$\pm$0.0412 & 0.67$\pm$0.0638 & 0.48$\pm$0.0771 \\ \hline
TGV                                                     & 31.84$\pm$3.9221 & 28.70$\pm$3.8181 & 25.52$\pm$3.6559 & 23.32$\pm$3.5221 & 0.93$\pm$0.0310 & 0.87$\pm$0.0580 & 0.75$\pm$0.1023 & 0.64$\pm$0.1372 \\ \hline
SOTV                                                    & 32.43$\pm$4.9447 & 29.39$\pm$4.5756 & 26.38$\pm$4.2877 & 24.41$\pm$4.0720 & 0.94$\pm$0.0308 & 0.89$\pm$0.0531 & 0.80$\pm$0.0902 & 0.71$\pm$0.1204 \\ \hline
Euler's elastica                                        & 32.12$\pm$4.0617 & 29.21$\pm$3.9857 & 26.30$\pm$3.8194 & 24.34$\pm$3.6208 & 0.94$\pm$0.0288 & 0.88$\pm$0.0526 & 0.78$\pm$0.0901 & 0.69$\pm$0.1208 \\ \hline
TWSO                                                    & 34.33$\pm$3.4049 & 31.12$\pm$3.0650 & 27.66$\pm$3.6540 & 25.26$\pm$3.2437 & 0.95$\pm$0.0204 & 0.90$\pm$0.0456 & 0.82$\pm$0.0853 & 0.73$\pm$0.1061 \\ \hline                         
\end{tabular}
} 
\label{tb:meanSD2}
\end{table}

\begin{figure}[h!] 
\vspace{-10pt}
\centering  
\subfigure[]{\includegraphics[width=0.32\textwidth]{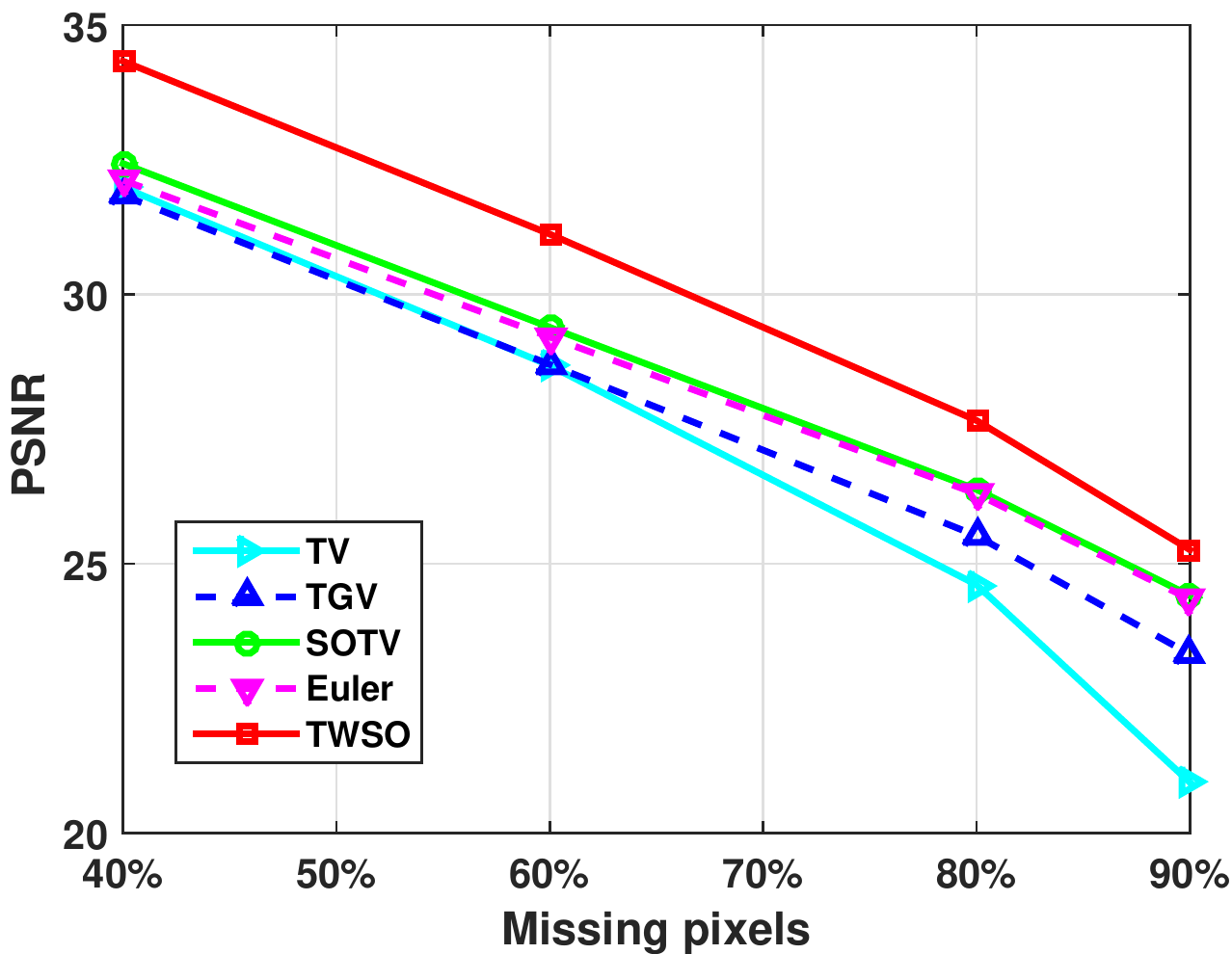}}
\subfigure[]{\includegraphics[width=0.32\textwidth]{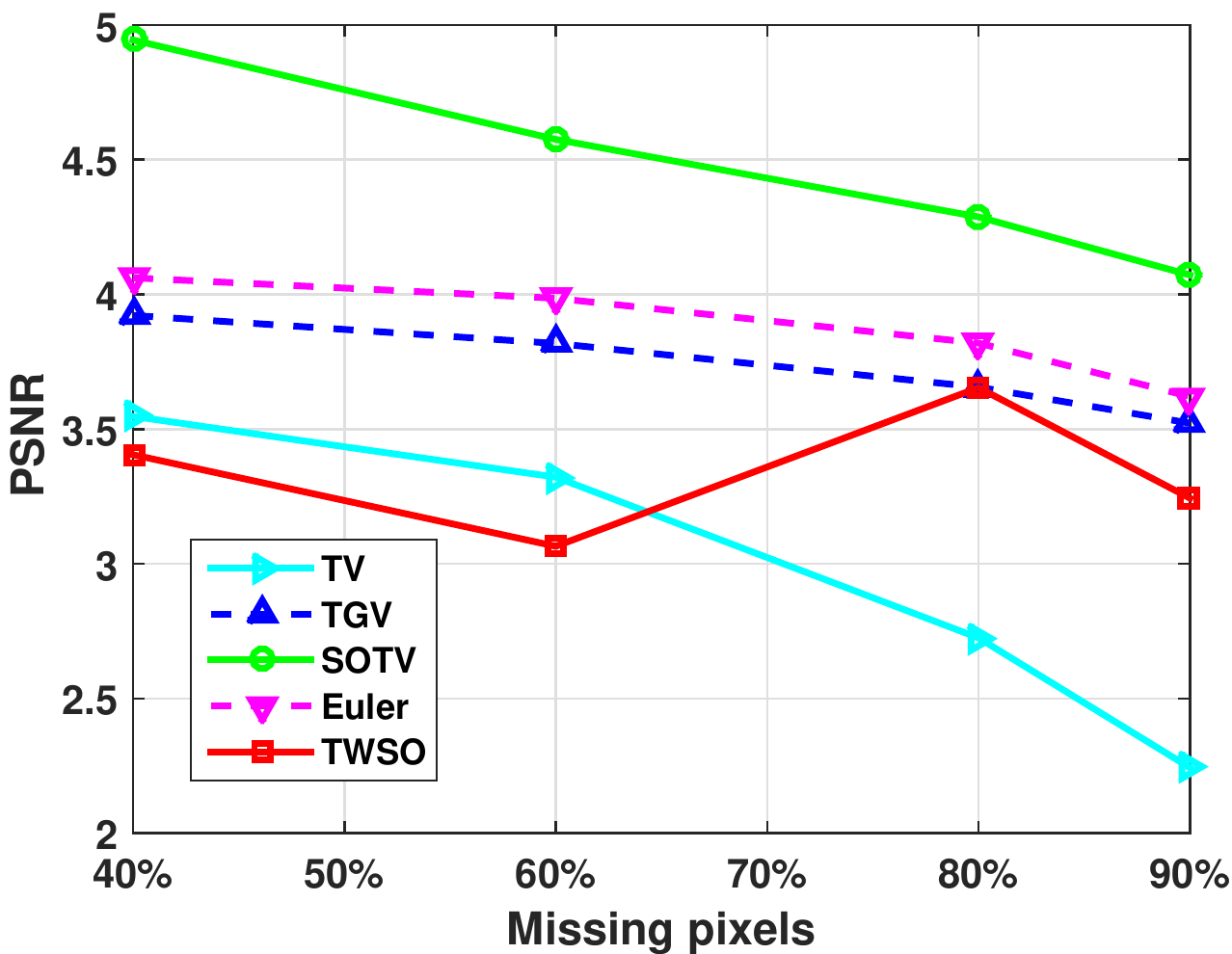}}\\
\vspace{-10pt}
\subfigure[]{\includegraphics[width=0.33\textwidth]{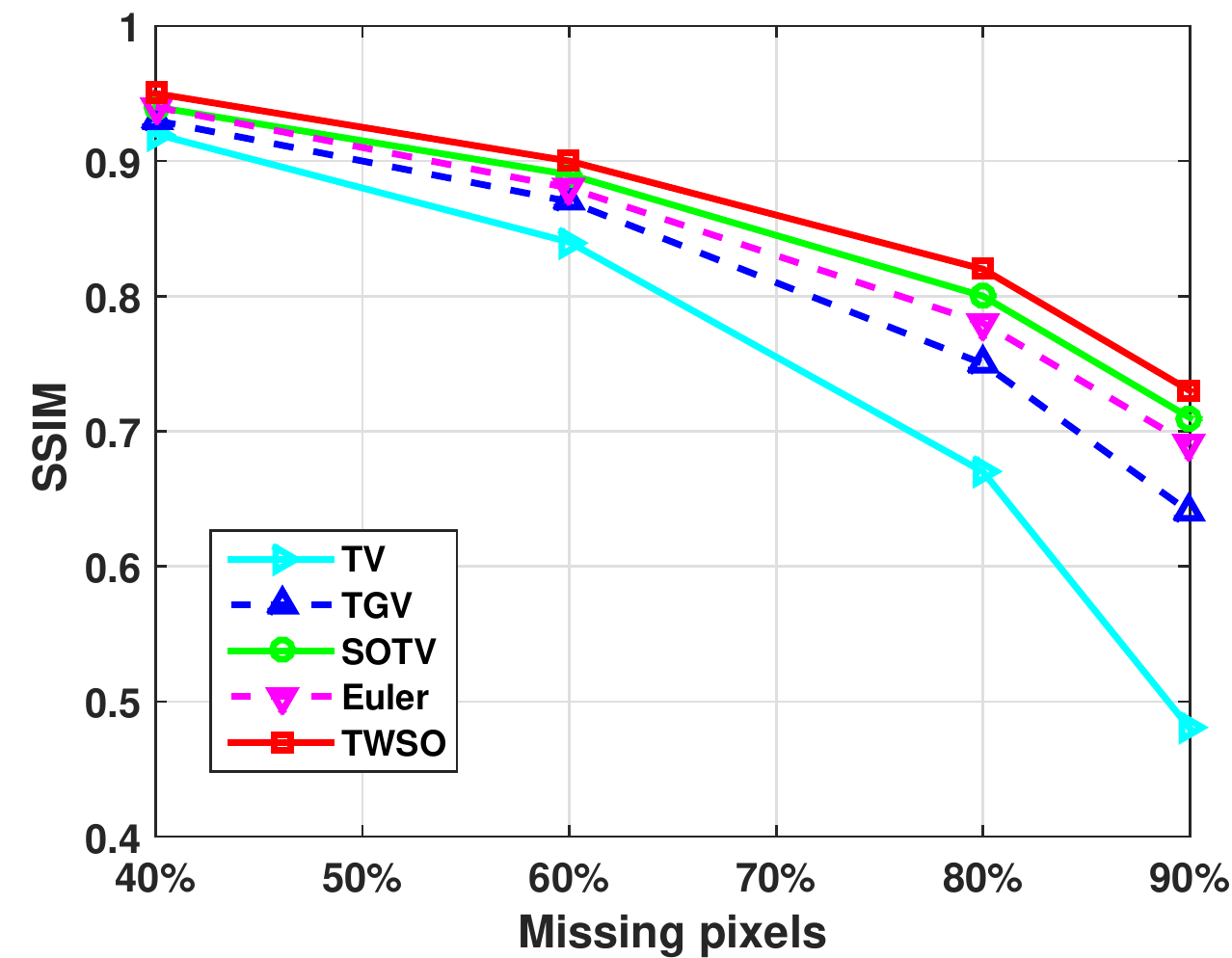}}
\subfigure[]{\includegraphics[width=0.325\textwidth]{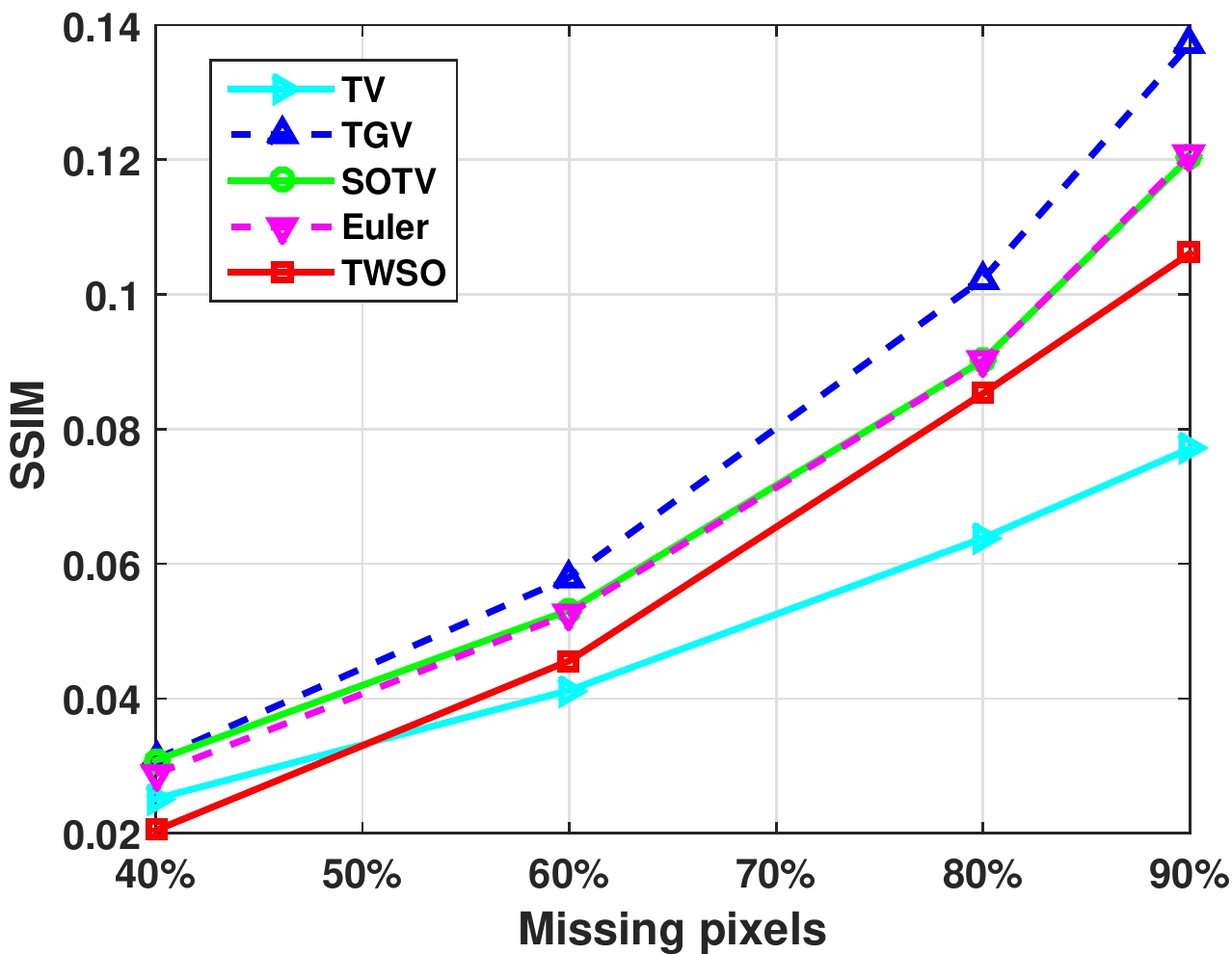}}
\vspace{-10pt}
\caption{Plots of mean and standard derivation in Table~\ref{tb:meanSD2} obtained by 5 different methods over 100 images from the Berkeley database BSDS500 with 4 different random masks.. (a) and (b): mean and standard derivation plots of PSNR; (c) and (d): mean and standard derivation plots of SSIM.}
\label{fig:meanSD2}
\end{figure}
Finally, we evaluate the TV, SOTV, TGV, Euler's elastica and TWSO on the Berkeley database BSDS500 for image inpainting. We use 4 random masks (i.e. 40$\%$, 60$\%$, 80$\%$ and 90$\%$ pixels are missing) for each of the 100 images randomly selected from the database. The performance of each method for each mask is measured in terms of mean and standard derivation of PSNR and SSIM over all 100 images. The results are demonstrated in the following Table~\ref{tb:meanSD2} and Figure~\ref{fig:meanSD2}. The accuracy of the inpainting results obtained by different methods decreases as the percentage of missing pixels region becomes larger. The highest averaged PSNR values are again achieved by the TWSO, demonstrating its effectiveness for image inpainting.    

\subsection{Image Denoising}
For denoising images corrubpted by Gaussian noise, we use $p=2$ in (\ref{eq:TWSO}) and the $\Gamma$ in (\ref{eq:TWSO}) is the same as the image domain $\Omega$. We compare the proposed TWSO model with the Perona-Malik (PM) \cite{perona1990scale}, coherent enhancing diffusion (CED) \cite{weickert1998anisotropic}, total variation (TV) \cite{rudin1992nonlinear}, second order total variation (SOTV) \cite{lysaker2003noise,bergounioux2010second,papafitsoros2013combined}, total generalised variation (TGV) \cite{bredies2010total}, and extended anisotropic diffusion model\footnote[4]{\tiny{Code: http://liu.diva-portal.org/smash/get/diva2:543914/SOFTWARE01.zip}} (EAD) \cite{aastrom2012tensor} on both synthetic and real images. 
\begin{figure}[h!] 
\centering  
{\includegraphics[width=0.103\textwidth]{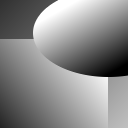}}
{\includegraphics[width=0.103\textwidth]{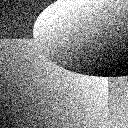}}
{\includegraphics[width=0.103\textwidth]{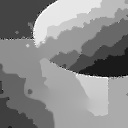}}
{\includegraphics[width=0.103\textwidth]{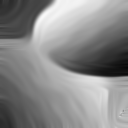}}
{\includegraphics[width=0.103\textwidth]{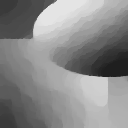}}
{\includegraphics[width=0.103\textwidth]{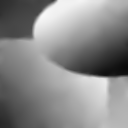}}
{\includegraphics[width=0.103\textwidth]{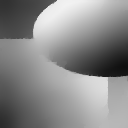}}
{\includegraphics[width=0.103\textwidth]{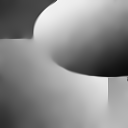}}
{\includegraphics[width=0.103\textwidth]{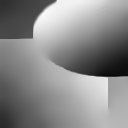}}\\
\vspace{-5pt}
\subfigure[]{\includegraphics[width=0.103\textwidth,height=0.103\textwidth]{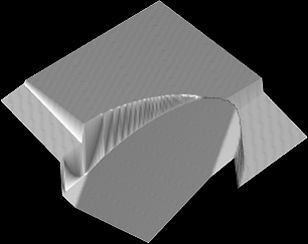}}
\subfigure[]{\includegraphics[width=0.103\textwidth,height=0.103\textwidth]{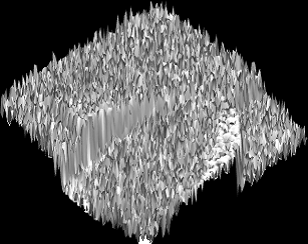}}
\subfigure[]{\includegraphics[width=0.103\textwidth,height=0.103\textwidth]{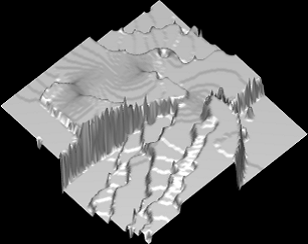}}
\subfigure[]{\includegraphics[width=0.103\textwidth,height=0.103\textwidth]{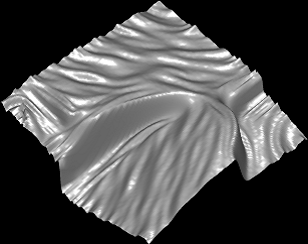}}
\subfigure[]{\includegraphics[width=0.103\textwidth,height=0.103\textwidth]{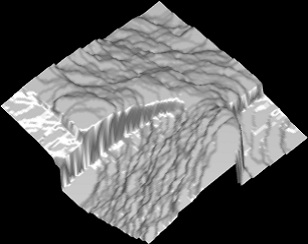}}
\subfigure[]{\includegraphics[width=0.103\textwidth,height=0.103\textwidth]{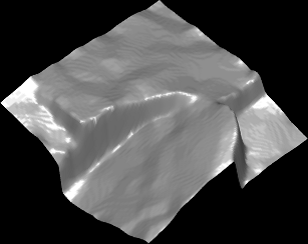}}
\subfigure[]{\includegraphics[width=0.103\textwidth,height=0.103\textwidth]{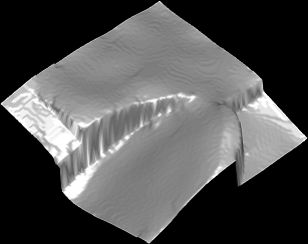}}
\subfigure[]{\includegraphics[width=0.103\textwidth,height=0.103\textwidth]{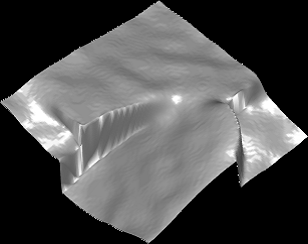}}
\subfigure[]{\includegraphics[width=0.103\textwidth,height=0.103\textwidth]{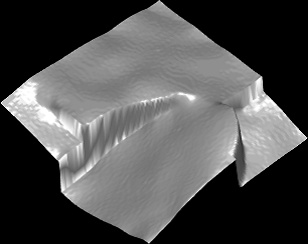}}\\
\vspace{-10pt}
\caption{Denoising results of a synthetic test image. (a): Clean data; (b): Image corrupted by 0.02 variance Gaussian noise; (c): PM result; (d): CED result; (e): TV result; (f): SOTV result; (g): TGV result; (h): EAD result; (i): TWSO result. The second row shows the isosurface rendering of the corresponding results above.}
\label{fig:denoise1}
\end{figure}

Figure~\ref{fig:denoise1} shows the denoised results on a synthetic image (a) by the different methods. The results by the PM and TV models, as shown in (c) and (e), have a jagged appearance (i.e. staircase artefact). However, the scale-space-based PM model shows much stronger staircase effect than the energy-based variational TV model for piecewise smooth image denoising. Due to the anisotropy, the result (d) by the CED method displays strong directional characteristics. Due to the high order derivatives involved, the SOTV, TGV and TWSO models can eliminate the staircase artefact. However, because the SOTV imposes too much regularity on the image, it smears the sharp edges of the objects in (f). Better results are obtained by the TGV, EAD and TWSO models, which neither produce the staircase artefact nor blur object edges, though TGV leaves some noise near the discontinuities and EAD over-smooths image edges, as shown in (g) and (h). 

\begin{figure}[h!] 
\centering  
{\includegraphics[width=0.103\textwidth]{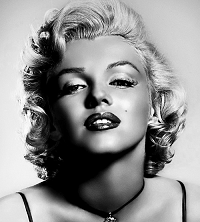}}
{\includegraphics[width=0.103\textwidth]{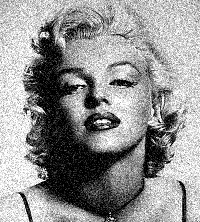}}
{\includegraphics[width=0.103\textwidth]{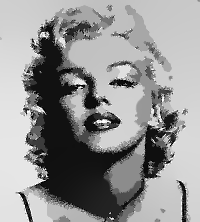}}
{\includegraphics[width=0.103\textwidth]{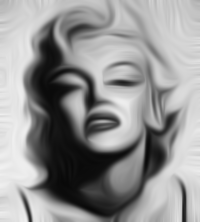}}
{\includegraphics[width=0.103\textwidth]{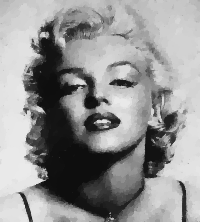}}
{\includegraphics[width=0.103\textwidth]{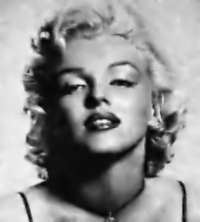}}
{\includegraphics[width=0.103\textwidth]{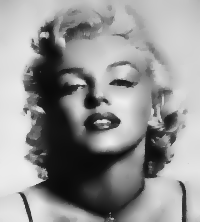}}
{\includegraphics[width=0.103\textwidth]{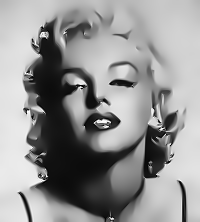}}
{\includegraphics[width=0.103\textwidth]{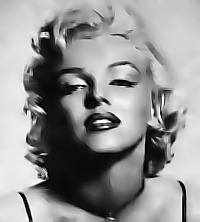}}\\
\vspace{-5pt}
\subfigure[]{\includegraphics[width=0.103\textwidth]{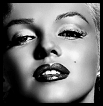}}
\subfigure[]{\includegraphics[width=0.103\textwidth]{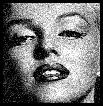}}
\subfigure[]{\includegraphics[width=0.103\textwidth]{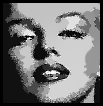}}
\subfigure[]{\includegraphics[width=0.103\textwidth]{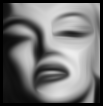}}
\subfigure[]{\includegraphics[width=0.103\textwidth]{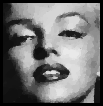}}
\subfigure[]{\includegraphics[width=0.103\textwidth]{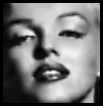}}
\subfigure[]{\includegraphics[width=0.103\textwidth]{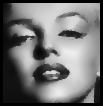}}
\subfigure[]{\includegraphics[width=0.103\textwidth]{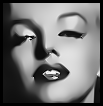}}
\subfigure[]{\includegraphics[width=0.103\textwidth]{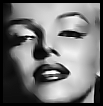}}\\
\vspace{-10pt}
\caption{Noise reduction results of a real test image. (a) Clean data; (b) Noisy data corrupted by 0.015 variance Gaussian noise; (c): PM result; (d): CED result; (e): TV result; (f): SOTV result; (g): TGV result; (h): EAD result; (i): TWSO result. The second row shows local magnification of the corresponding results in the first row.}
\label{fig:denoise2}
\vspace{-10pt}
\end{figure}

Figure~\ref{fig:denoise2} presents the denoised results on a real image (a) by the different methods. Both the CED and the proposed TWSO models use the anisotropic diffusion tensor $\textbf{T}$. CED distorts the image while TWSO does not. The reason for this is that TWSO uses the eigenvalues of $\textbf{T}$ defined in (\ref{eq:edgePresEigVal}), which has two advantages: i) it allows us to control the degree of the anisotropy of TWSO. Specifically, if the contrast parameter $C$ in (\ref{eq:edgePresEigVal}) goes to infinity, the TWSO model degenerates to the pure isotropic SOTV model. The larger $C$ is, the less anisotropy TWSO has. However, the eigenvalues (\ref{eq:struPresEigVal}) used in CED are not able to adjust the anisotropy; ii) it can decide if there exists diffusion in TWSO along the direction parallel to the image gradient direction. The diffusion halts along the image gradient direction if $\lambda_1$ in (\ref{eq:edgePresEigVal}) is small. The diffusion is however encouraged if $\lambda_1$ is large. The eigenvalue $\lambda_1$ used in (\ref{eq:struPresEigVal}) however remains small (i.e. $\lambda_1\ll 1$) all the time, meaning that the diffusion in CED is always prohibited. CED thus only prefers the orientation that is perpendicular to the image gradient. This explains why CED distorts the image. In fact, by choosing a suitable $C$, (\ref{eq:edgePresEigVal}) allows TWSO to diffuse the noise along object edges and prohibit the diffusion across edges without showing any orientations.

\begin{table}[h]
\caption{Comparison of PSNR and SSIM using different methods on Figure~\ref{fig:denoise1} and~\ref{fig:denoise2} with different noise variances.}
\vspace{-10pt}
\resizebox{\columnwidth}{!}{
\begin{tabular}{|c||c|c|c|c|c|c|c|c|c|c||c|c|c|c|c|c|c|c|c|c|}
\hline
														& \multicolumn{10}{c||}{Figure~\ref{fig:denoise1}}    & \multicolumn{10}{c|}{Figure~\ref{fig:denoise2}}    \\  \cline{2-21}                       
                                                        & \multicolumn{5}{c|}{PSNR value}   & \multicolumn{5}{c||}{SSIM value}  & \multicolumn{5}{c|}{PSNR value}   & \multicolumn{5}{c|}{SSIM value} \\    \hline                        
Noise variance                                          & 0.005   & 0.01    & 0.015   & 0.02    & 0.025   & 0.005  & 0.01      & 0.015     & 0.02     & 0.025     & 0.005   & 0.01       & 0.015      & 0.02       & 0.025     & 0.005    & 0.01     & 0.015    & 0.02     & 0.025  \\ \hline
Degraded                                                & 23.1328 & 20.2140 & 18.4937 & 17.3647 & 16.4172 & 0.2776 & 0.1827    & 0.1419    & 0.1212    & 0.1038   & 23.5080 & 20.7039    & 19.0771    & 17.9694    & 17.1328   & 0.5125   & 0.4167   & 0.3625   & 0.3261   & 0.3009 \\ \hline
PM                                                      & 26.7680 & 23.3180 & 20.0096 & 19.4573 & 18.9757 & 0.8060 & 0.7907    & 0.7901    & 0.7850    & 0.7822   & 25.8799 & 24.0247    & 22.4998    & 22.0613    & 21.1334   & 0.7819   & 0.7273   & 0.6991   & 0.6760   & 0.6579 \\ \hline
CED                                                     & 29.4501 & 29.2004 & 28.4025 & 28.2601 & 27.7806 & 0.9414 & 0.9264    & 0.9055    & 0.8949    & 0.8858   & 23.8593 & 21.3432    & 20.2557    & 20.1251    & 20.0243   & 0.6728   & 0.6588   & 0.6425   & 0.6375   & 0.6233 \\ \hline
TV                                                      & 32.6027 & 29.9910 & 28.9663 & 28.3594 & 27.1481 & 0.9387 & 0.9233    & 0.9114    & 0.9017    & 0.8918   & 27.3892 & 25.5942    & 24.4370    & 23.5511    & 22.8590   & 0.8266   & 0.7826   & 0.7529   & 0.7331   & 0.7149 \\ \hline
SOTV                                                    & 32.7415 & 31.1342 & 30.1882 & 29.4404 & 28.3784 & 0.9653 & 0.9552    & 0.9474    & 0.9444    & 0.9332   & 26.1554 & 24.6041    & 23.5738    & 23.2721    & 22.5986   & 0.8252   & 0.7841   & 0.7546   & 0.7367   & 0.7201 \\ \hline
TGV                                                     & 36.1762 & 34.6838 & 33.1557 & 32.0843 & 30.6564 & 0.9821 & 0.9745    & 0.9688    & 0.9644    & 0.9571   & 27.5006 & 25.6966    & 24.5451    & 23.6456    & 22.9540   & 0.8375   & 0.7937   & 0.7659   & 0.7456   & 0.7386 \\ \hline
EAD                                                     & 34.8756 & 33.7161 & 32.6091 & 31.6987 & 30.6599 & 0.9764 & 0.9719    & 0.9678    & 0.9610    & 0.9548   & 27.0495 & 25.0156    & 24.1903    & 23.4999    & 23.2524   & 0.8232   & 0.7791   & 0.7606   & 0.7424   & 0.7382 \\ \hline
TWSO                                                    & \textbf{36.3192} &\textbf{34.7220} & \textbf{33.4009} & \textbf{32.7334} & \textbf{31.5537} &\textbf{0.9855} & \textbf{0.9771}    & \textbf{0.9726}    & \textbf{0.9704}    & \textbf{0.9658}   &\textbf{27.5997} & \textbf{25.8511}    & \textbf{24.9060}    &\textbf{24.1328}    &\textbf{23.4983}  & \textbf{0.8437}   & \textbf{0.8063}   & \textbf{0.7811}   &\textbf{0.7603}   & \textbf{0.7467} \\ 
\hline                               
\end{tabular}
} 
\label{tb:4metrics}
\end{table}

\begin{figure}[h!] 
\vspace{-10pt}
\centering  
\subfigure[]{\includegraphics[width=0.23\textwidth]{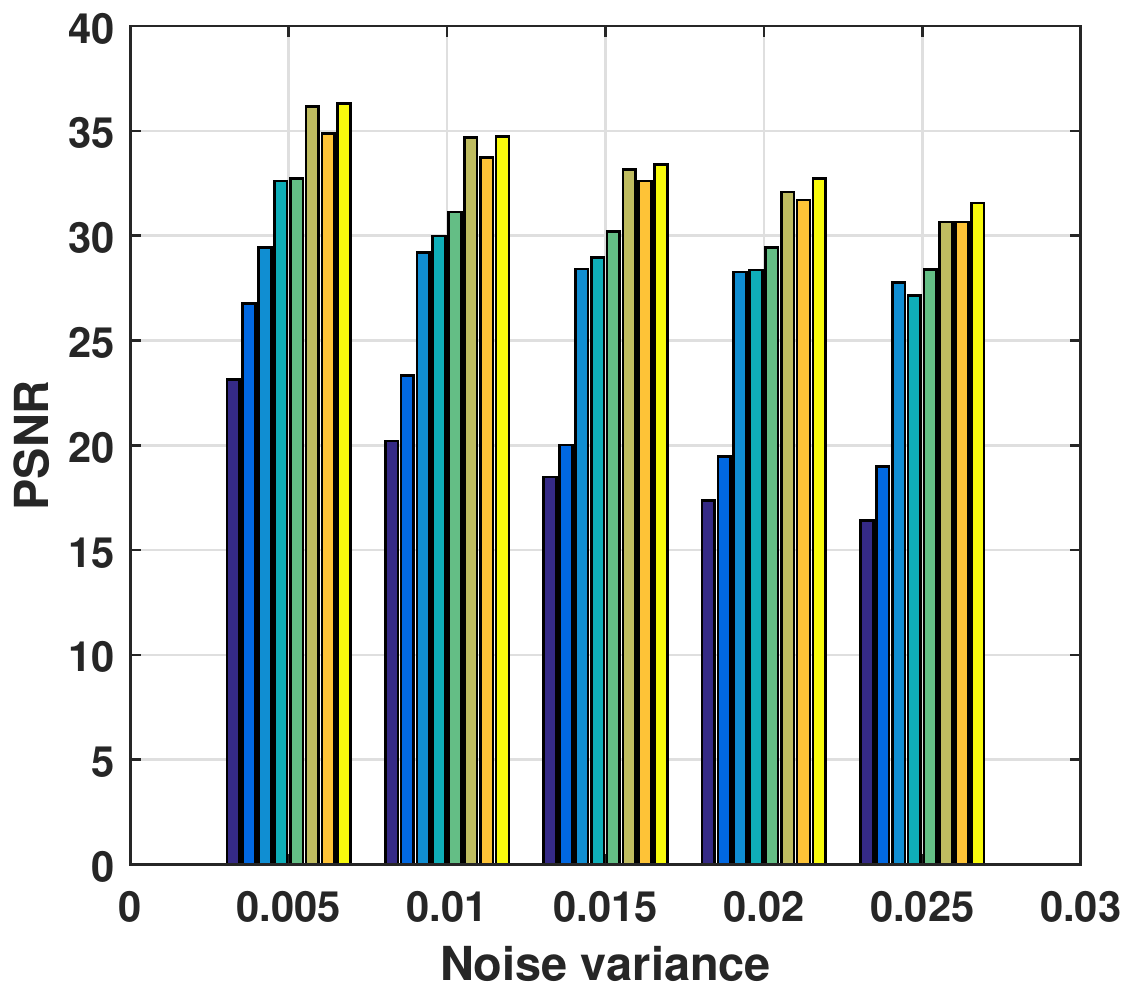}}
\subfigure[]{\includegraphics[width=0.23\textwidth]{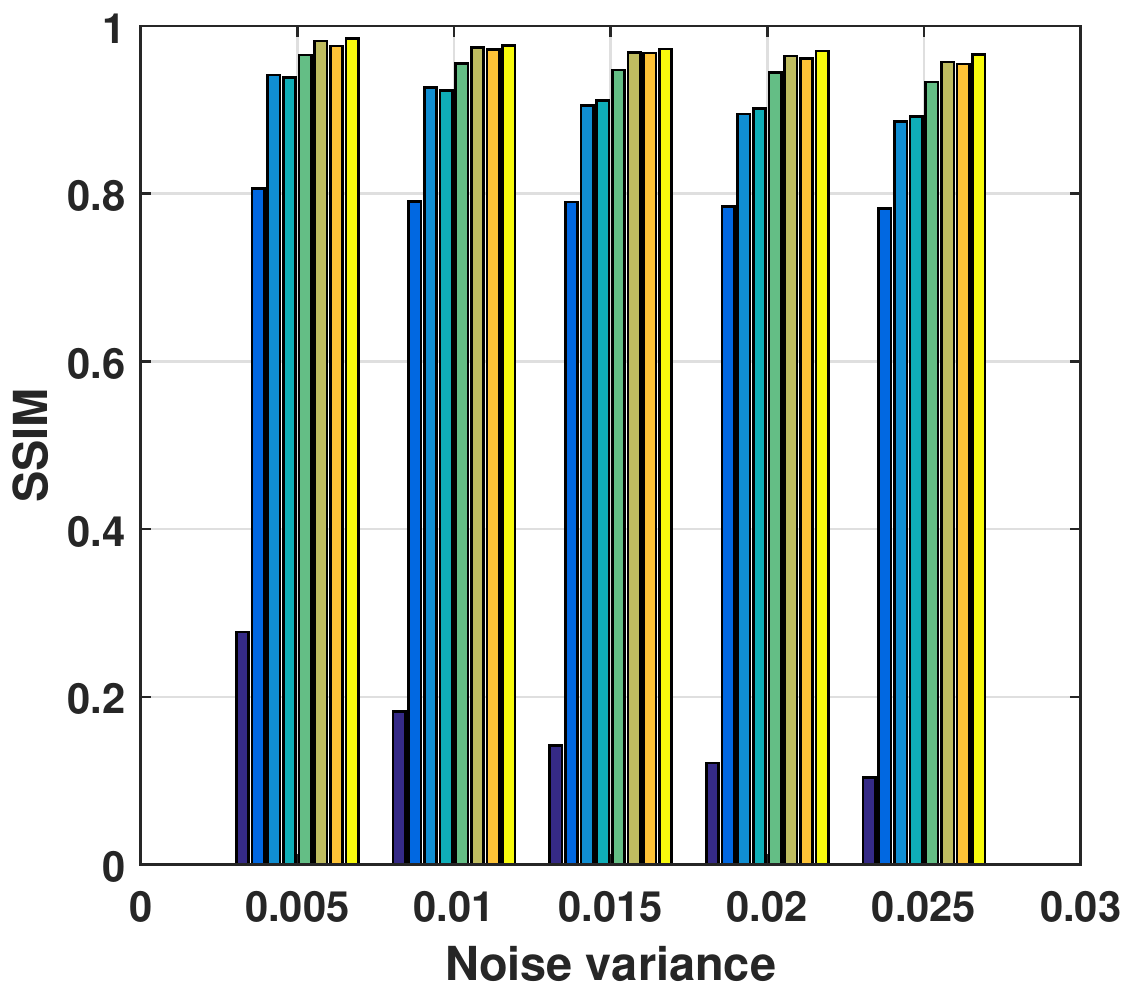}}
\subfigure[]{\includegraphics[width=0.23\textwidth]{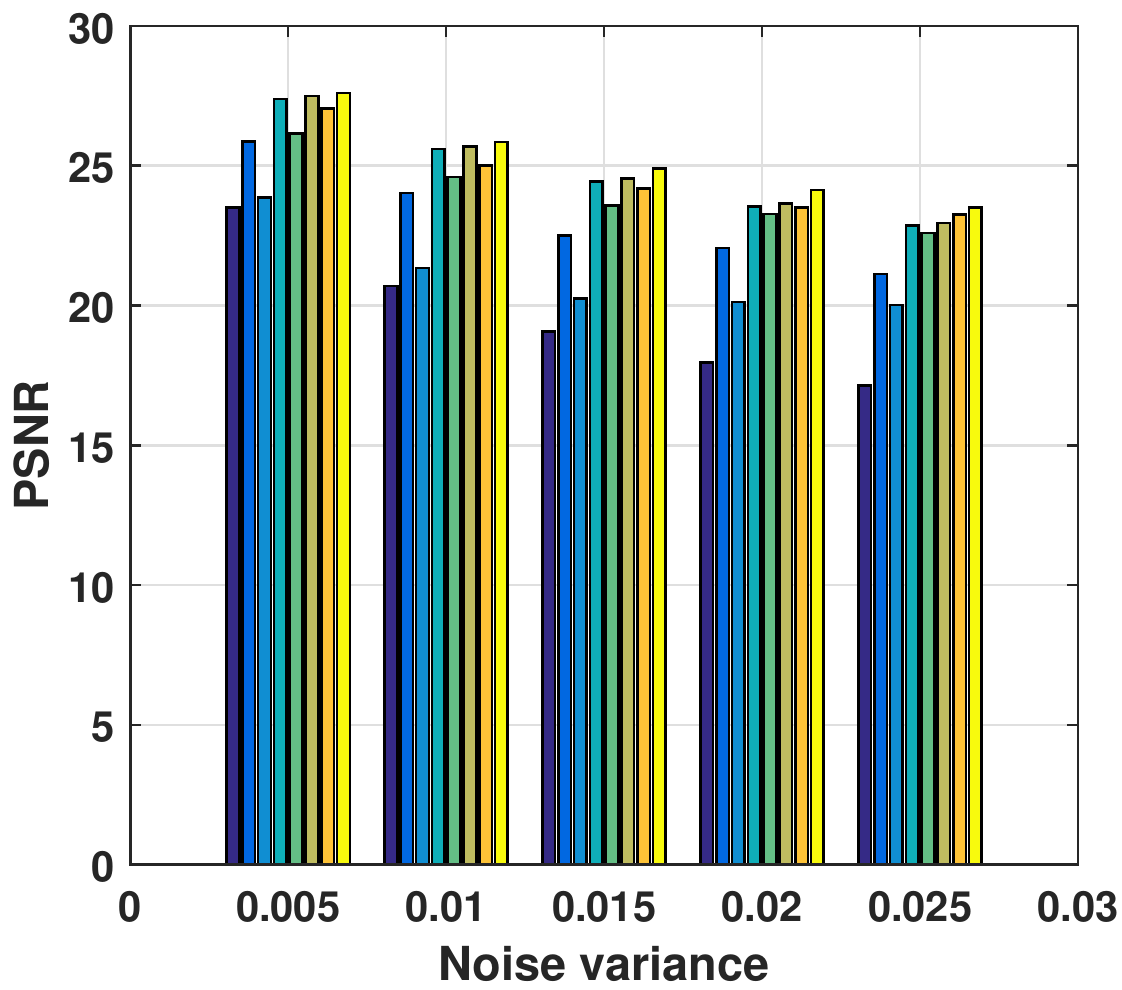}}
\subfigure[]{\includegraphics[width=0.23\textwidth]{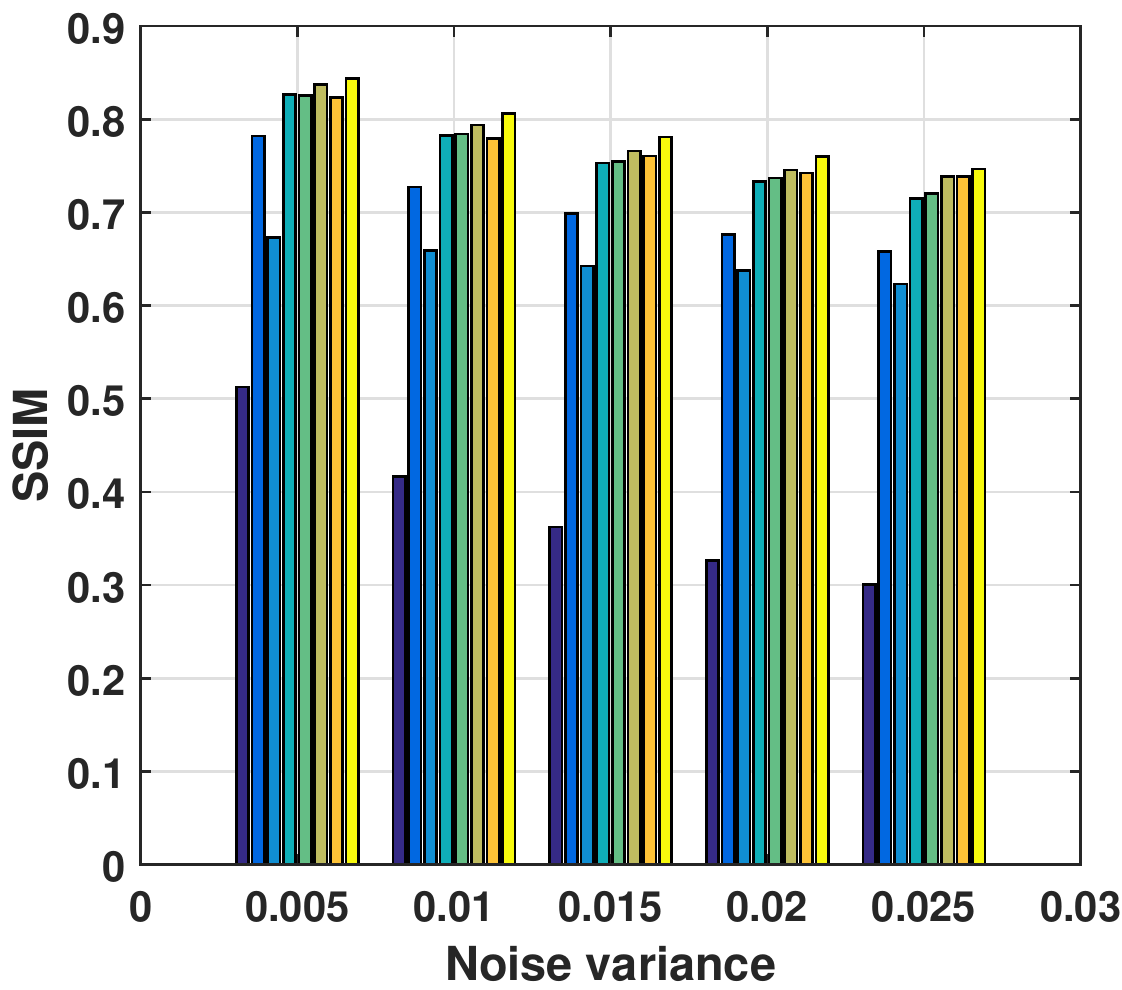}}
\subfigure  {\includegraphics[width=0.032\textwidth]{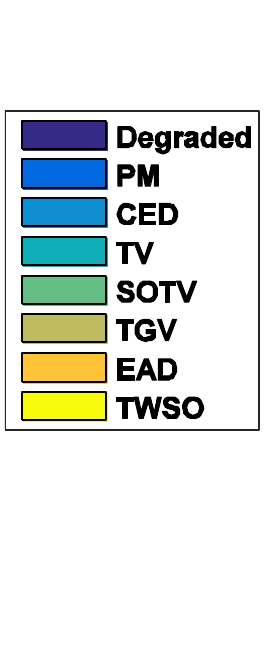}}
\vspace{-10pt}
\caption{Quantitative image quality evaluation for Gaussian noise reduction in Table~\ref{tb:4metrics}. (a) and (b): PSNR and SSIM values of various denoised methods for Figure~\ref{fig:denoise1} (a) corrupted by different noise levels; (c) and (d): PSNR and SSIM values of various denoised methods for Figure~\ref{fig:denoise2} (a) corrupted by different noise levels.}
\label{fig:denoise21}
\end{figure}

In addition to qualitative evaluation of the different methods in Figure~\ref{fig:denoise1} and~\ref{fig:denoise2}, we also calculate the PSNR and SSIM values for these methods and show them in Table~\ref{tb:4metrics} and Figure~\ref{fig:denoise21}. These metrics show that the PDE-based methods (i.e. PM and CED) perform worse than the variational methods (i.e. TV, SOTV, TGV, EAD and TWSO). The TV model introduces staircase effect, SOTV blurs the image edges, and EAD tends to over-smooth the image structure. The proposed TWSO model performs well in all the cases, validating the effectiveness of it for image denoising. 

\begin{figure}[h!] 
\vspace{-10pt}
\centering  
\subfigure[]{\includegraphics[width=0.32\textwidth]{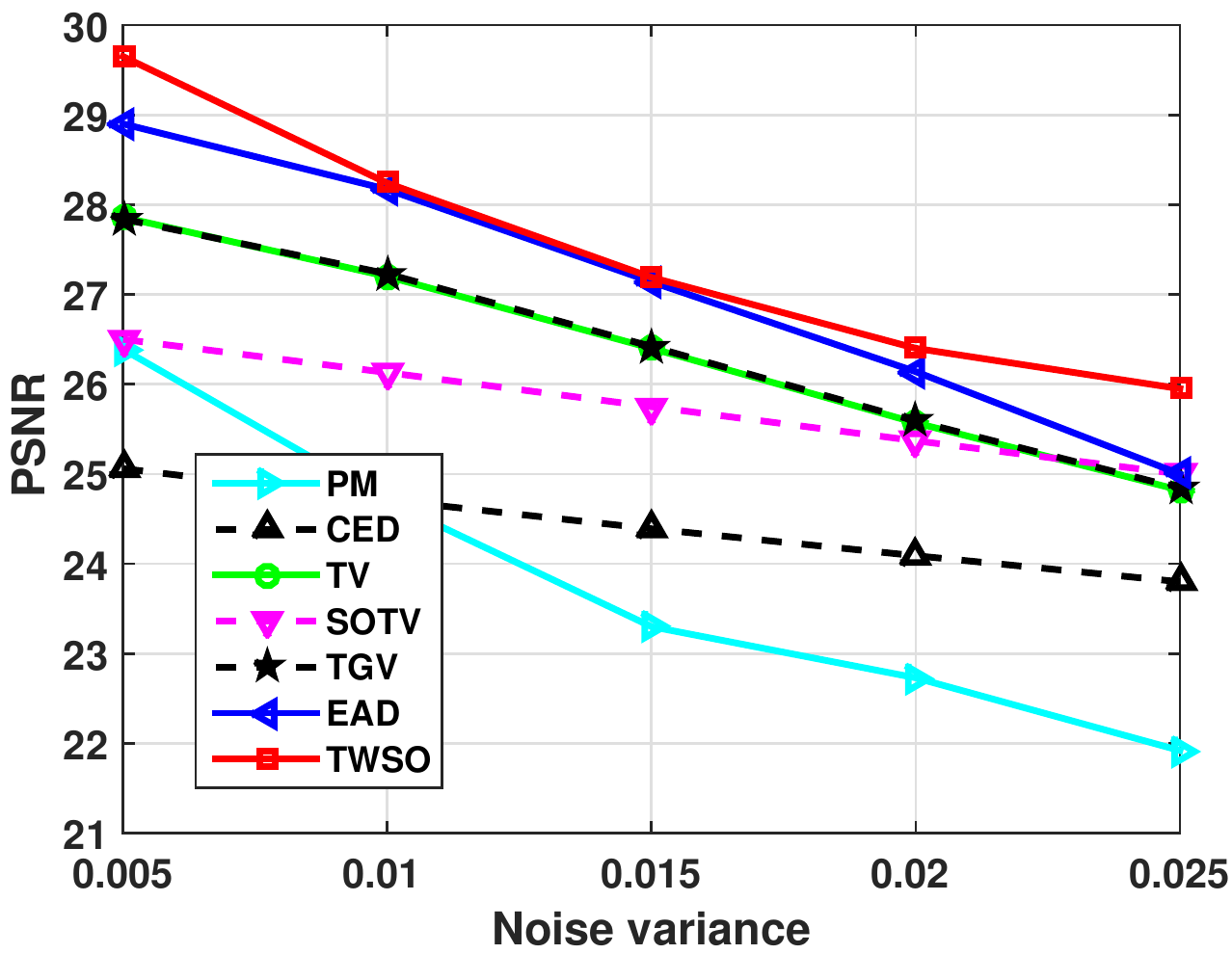}}
\subfigure[]{\includegraphics[width=0.32\textwidth]{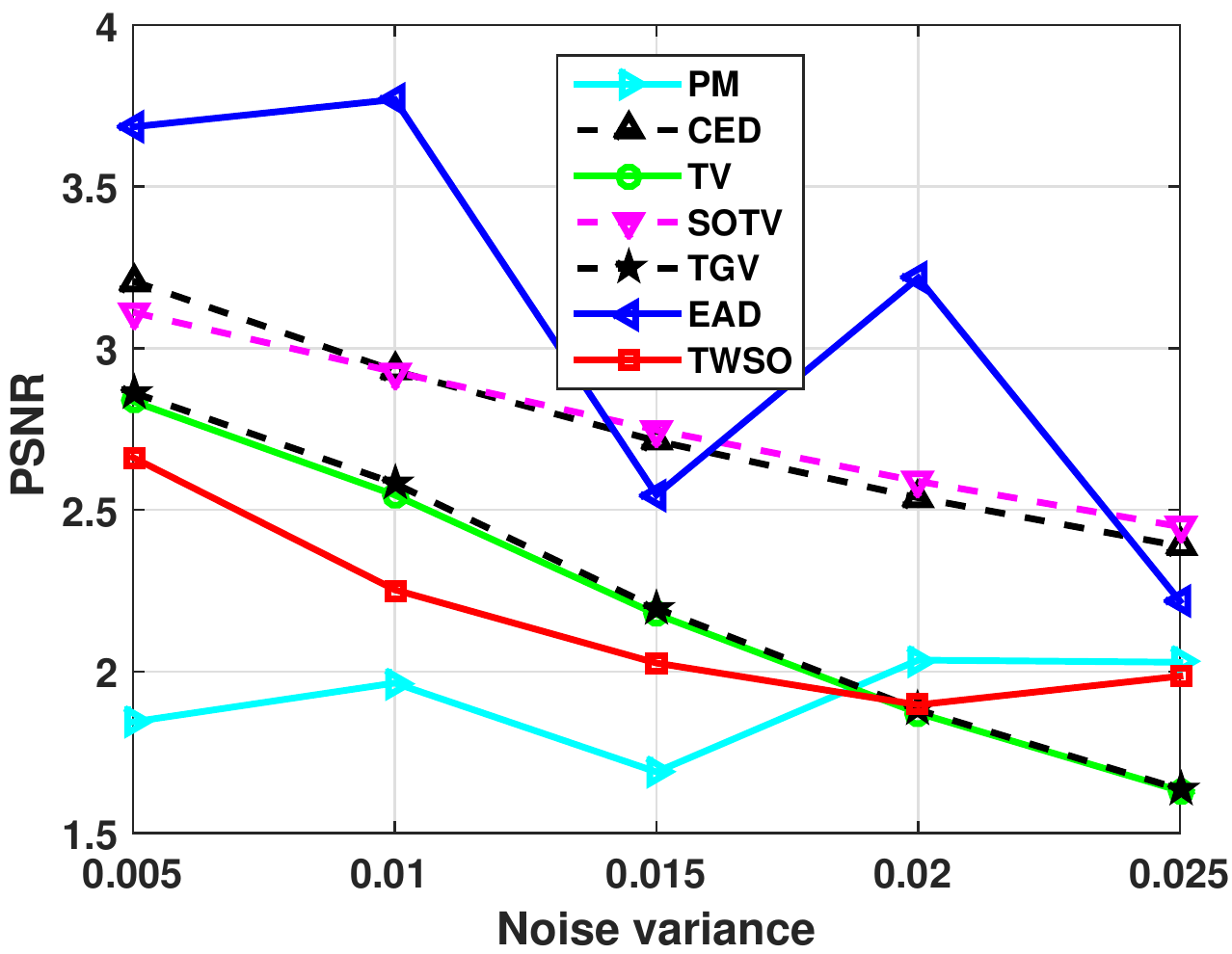}}\\
\vspace{-10pt}
\subfigure[]{\includegraphics[width=0.33\textwidth]{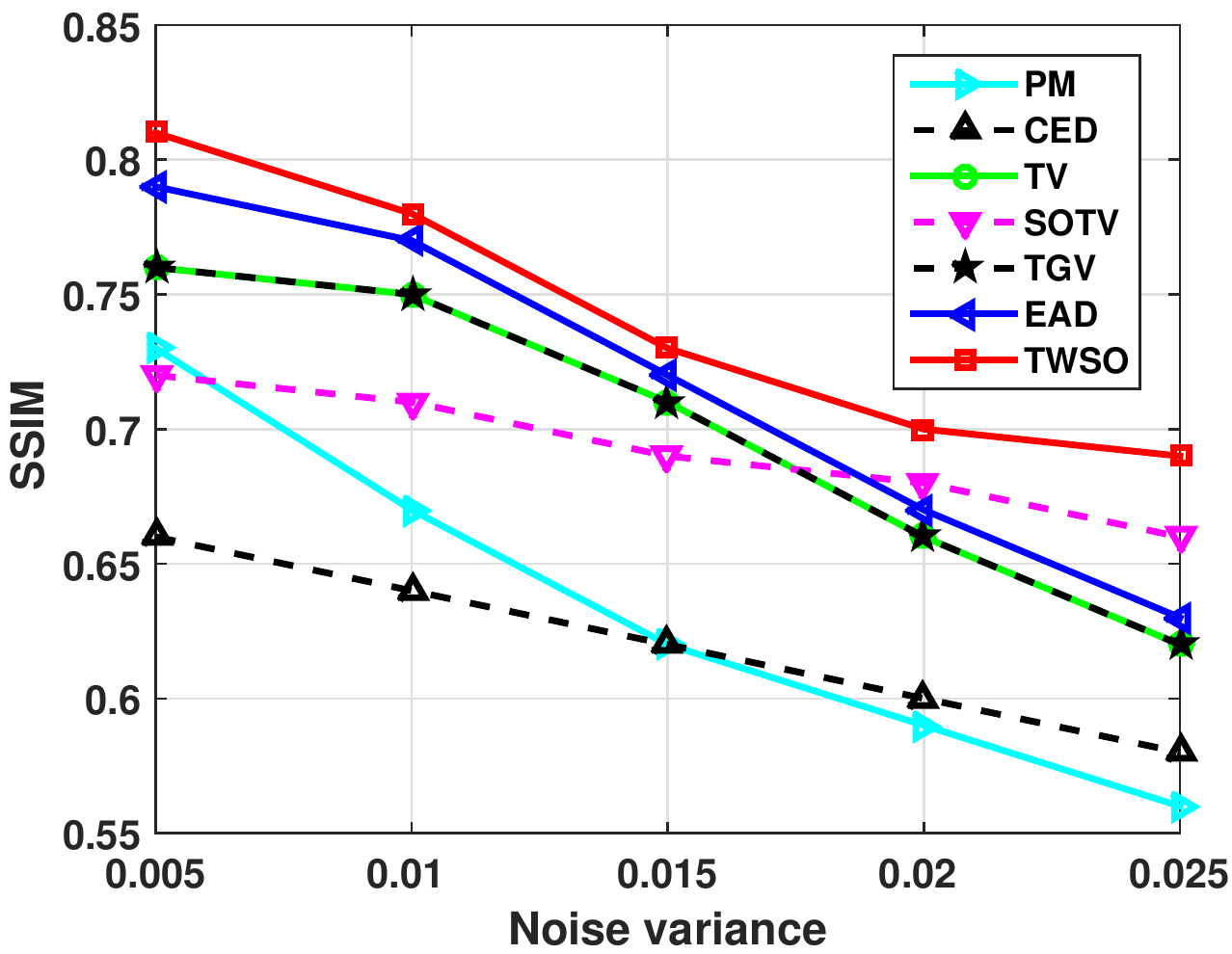}}
\subfigure[]{\includegraphics[width=0.32\textwidth]{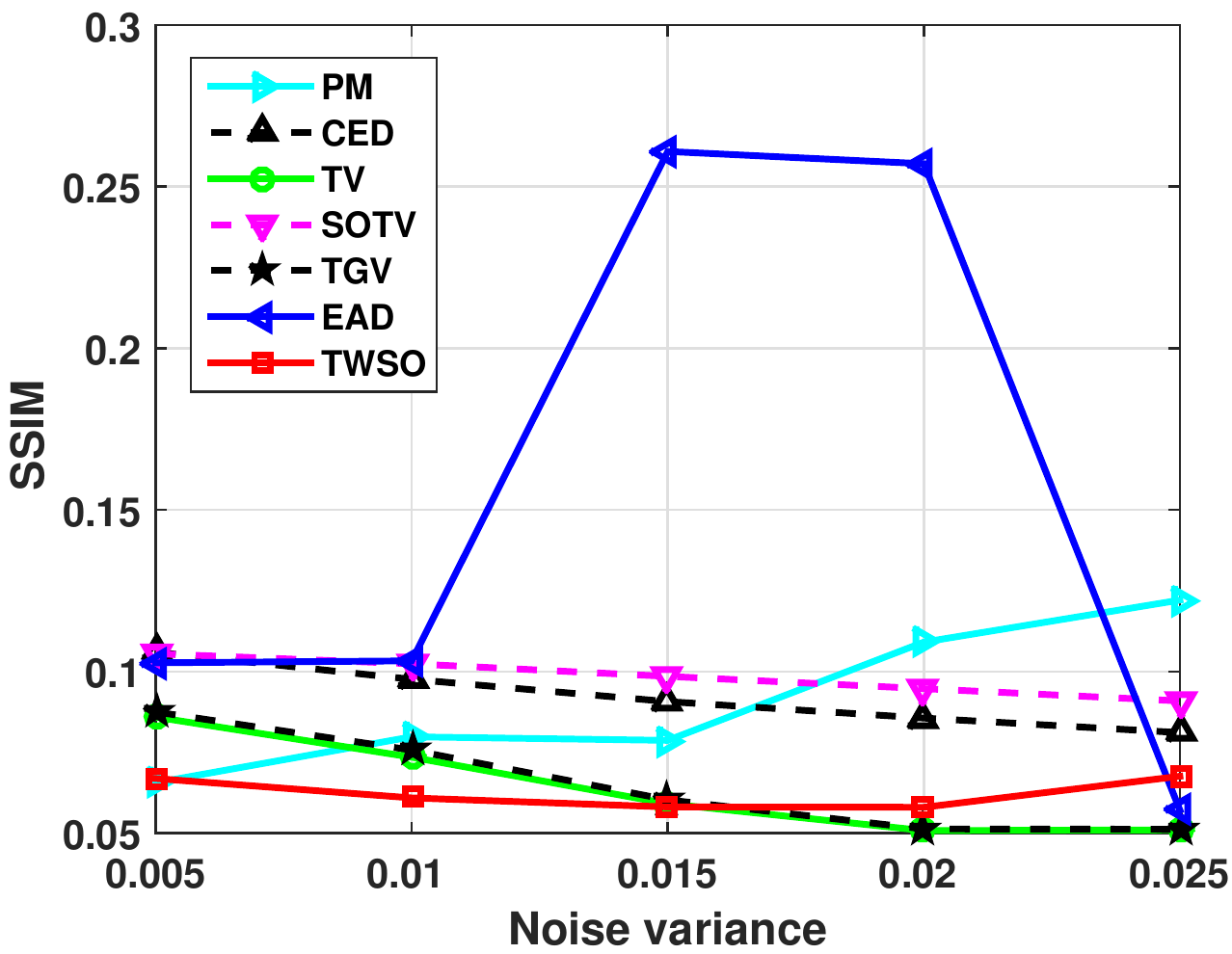}}
\vspace{-10pt}
\caption{Plots of mean and standard derivation in Table~\ref{tb:meanSD1} obtained by 7 different methods over 100 images from the Berkeley database BSDS500 with 5 different noise variances. (a) and (b): mean and standard derivation plots of PSNR; (c) and (d): mean and standard derivation plots of SSIM.}
\label{fig:meanSD1}
\end{figure}

We now evaluate the PM, CED, TV, SOTV, TGV, EAD and TWSO on the Berkeley database BSDS500 for image denoising. 100 images are first randomly selected from the database and each of the chosen images is corrupted by Gaussian noise with zero-mean and 5 variances ranging from 0.005 to 0.025 at 0.005 interval. The performance of each method for each noise variance is measured in terms of mean and standard derivation of PSNR and SSIM over all the 100 images. The final quantitative results are shown in Table~\ref{tb:meanSD1} and Figure~\ref{fig:meanSD1}. The mean values of PSNR and SSIM obtained by the TWSO remain the largest in all the cases. The standard derivation of PSNR and SSIM are smaller and relatively stable, indicating that the proposed TWSO is robust against the increasing level of noise and performs the best among all the 7 methods compared for image denoising. 

\begin{table}[h]
\caption{Mean and standard deviation (SD) of PSNR and SSIM calculated using 7 different methods for image denoising over 100 images from the Berkeley database BSDS500 with 5 different noise variances.}
\vspace{-10pt}
\resizebox{\columnwidth}{!}{
\begin{tabular}{|c||c|c|c|c|c||c|c|c|c|c|}
\hline
                                                        & \multicolumn{5}{c||}{PSNR value (Mean$\pm$ SD)}   & \multicolumn{5}{c|}{SSIM value (Mean$\pm$ SD)} \\ \hline \hline                         
Noise variance                                          & 0.005   & 0.01    & 0.015   & 0.02    & 0.025   & 0.005      & 0.01       & 0.015      & 0.02       & 0.025      \\ \hline
Degraded                                                & 23.19$\pm$0.2173 & 20.28$\pm$0.2831 & 18.61$\pm$0.3203 & 17.44$\pm$0.3492 & 16.55$\pm$0.3683 & 0.49$\pm$0.1229  & 0.37$\pm$0.1164 & 0.31$\pm$0.1078 & 0.27$\pm$0.1004 & 0.24$\pm$0.0941 \\ \hline
PM                                                      & 26.39$\pm$1.8442 & 24.70$\pm$1.9644 & 23.30$\pm$1.6893 & 22.73$\pm$2.0353 & 21.92$\pm$2.0287 & 0.73$\pm$0.0656  & 0.67$\pm$0.0798 & 0.62$\pm$0.0787 & 0.59$\pm$0.1091 & 0.56$\pm$0.1221 \\ \hline
CED                                                     & 25.06$\pm$3.2078 & 24.71$\pm$2.9300 & 24.39$\pm$2.7133 & 24.09$\pm$2.5376 & 23.80$\pm$2.3884 & 0.66$\pm$0.1060  & 0.64$\pm$0.0976 & 0.62$\pm$0.0907 & 0.60$\pm$0.0856 & 0.58$\pm$0.0811 \\ \hline
TV                                                      & 27.86$\pm$2.8359 & 27.20$\pm$2.5462 & 26.40$\pm$2.1771 & 25.57$\pm$1.8715 & 24.82$\pm$1.6308 & 0.76$\pm$0.0859  & 0.75$\pm$0.0735 & 0.71$\pm$0.0591 & 0.66$\pm$0.0508 & 0.62$\pm$0.0510 \\ \hline
SOTV                                                    & 26.50$\pm$3.1114 & 26.13$\pm$2.9269 & 25.75$\pm$2.7474 & 25.37$\pm$2.5885 & 25.00$\pm$2.4484 & 0.72$\pm$0.1055  & 0.71$\pm$0.1024 & 0.69$\pm$0.0985 & 0.68$\pm$0.0947 & 0.66$\pm$0.0909 \\ \hline
TGV                                                     & 27.84$\pm$2.8632 & 27.23$\pm$2.5818 & 26.42$\pm$2.1958 & 25.59$\pm$1.8814 & 24.84$\pm$1.6360 & 0.76$\pm$0.0875  & 0.75$\pm$0.0757 & 0.71$\pm$0.0603 & 0.66$\pm$0.0513 & 0.62$\pm$0.0512 \\ \hline
EAD                                                     & 28.90$\pm$3.6852 & 28.17$\pm$3.7699 & 27.14$\pm$2.5467 & 26.14$\pm$3.2178 & 25.00$\pm$2.2179 & 0.79$\pm$0.1027  & 0.77$\pm$0.1033 & 0.72$\pm$0.2609 & 0.67$\pm$0.2571 & 0.63$\pm$0.0571 \\ \hline
TWSO                                                    & 29.65$\pm$2.6622 & 28.24$\pm$2.2530 & 27.19$\pm$2.0262 & 26.40$\pm$1.8969 & 25.95$\pm$1.9856 & 0.81$\pm$0.0669  & 0.78$\pm$0.0609 & 0.73$\pm$0.0581 & 0.70$\pm$0.0580 & 0.69$\pm$0.0676 \\ 
\hline                               
\end{tabular}
} 
\label{tb:meanSD1}
\end{table}

\begin{figure}[h!] 
\vspace{-5pt}
\centering  
\subfigure{\includegraphics[width=0.14\textwidth]{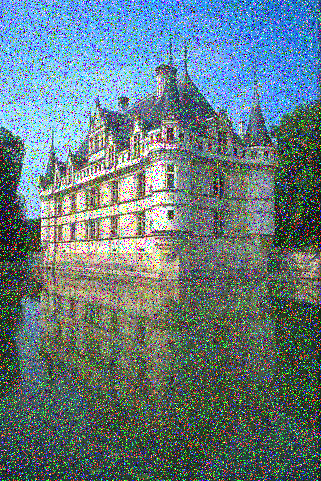}}
\subfigure{\includegraphics[width=0.14\textwidth]{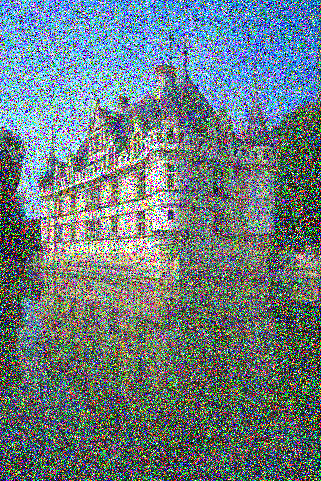}}
\subfigure{\includegraphics[width=0.14\textwidth]{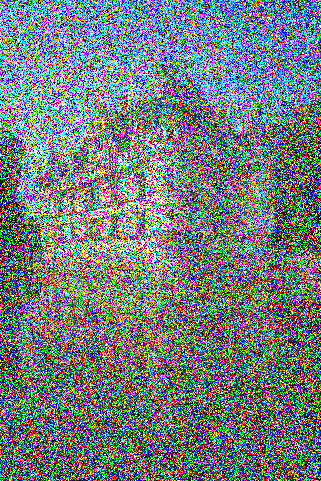}}
\subfigure{\includegraphics[width=0.14\textwidth]{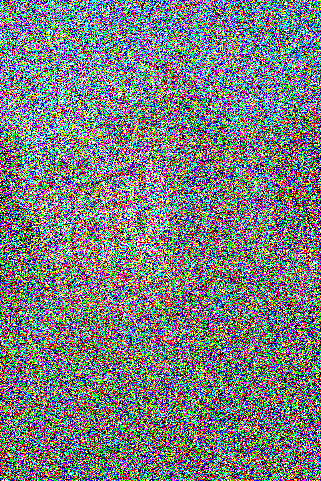}}
\subfigure{\includegraphics[width=0.14\textwidth]{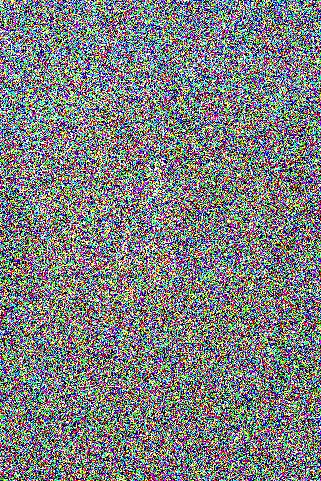}}\\
\vspace{-10pt}
\subfigure{\includegraphics[width=0.14\textwidth]{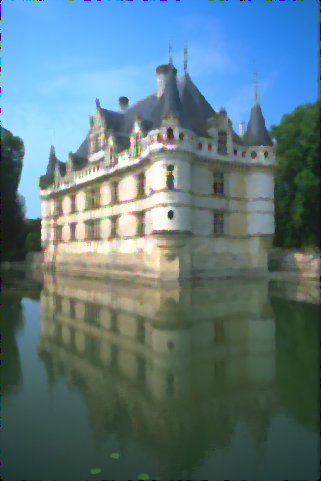}}
\subfigure{\includegraphics[width=0.14\textwidth]{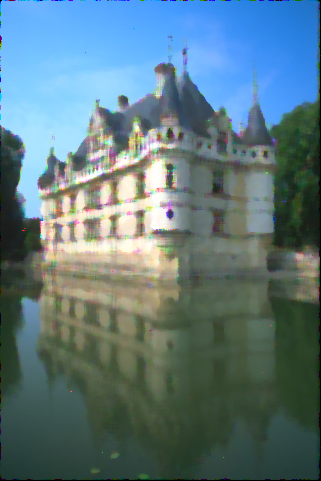}}
\subfigure{\includegraphics[width=0.14\textwidth]{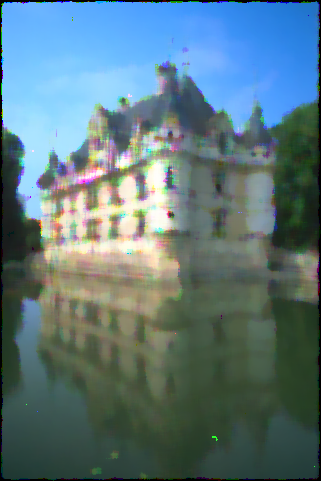}}
\subfigure{\includegraphics[width=0.14\textwidth]{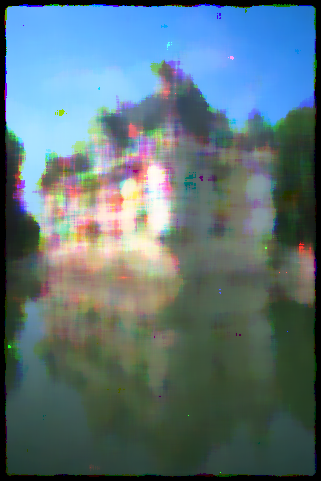}}
\subfigure{\includegraphics[width=0.14\textwidth]{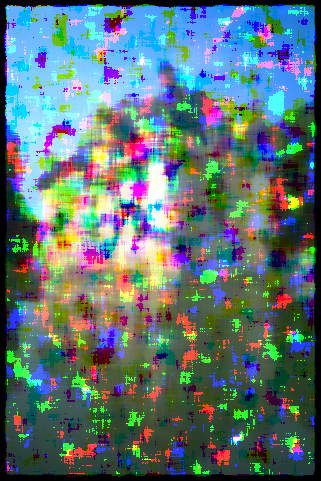}}\\
\vspace{-10pt}
\subfigure{\includegraphics[width=0.14\textwidth]{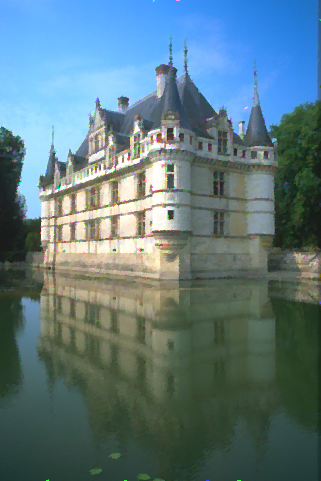}}
\subfigure{\includegraphics[width=0.14\textwidth]{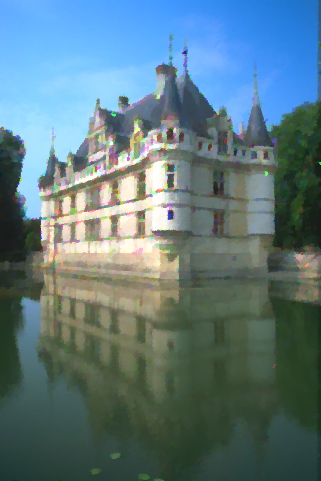}}
\subfigure{\includegraphics[width=0.14\textwidth]{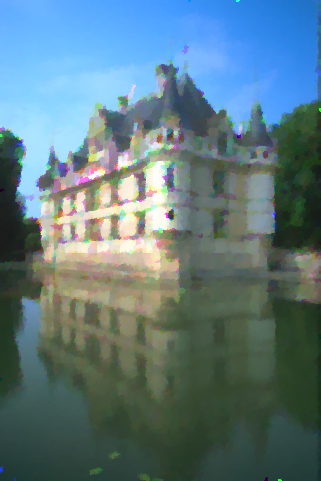}}
\subfigure{\includegraphics[width=0.14\textwidth]{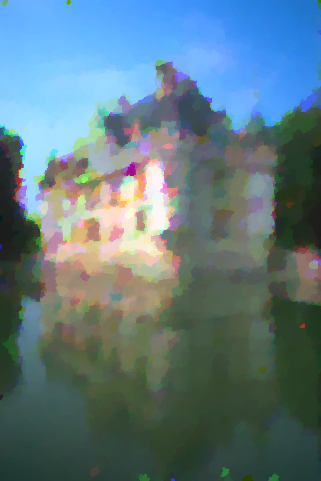}}
\subfigure{\includegraphics[width=0.14\textwidth]{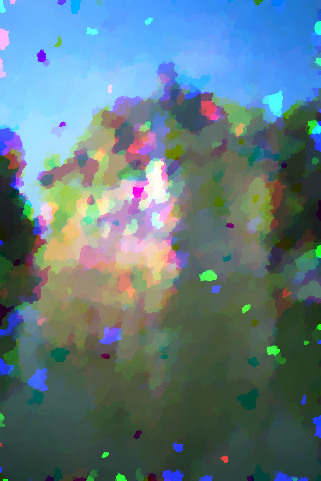}}\\
\vspace{-10pt}
\subfigure{\includegraphics[width=0.14\textwidth]{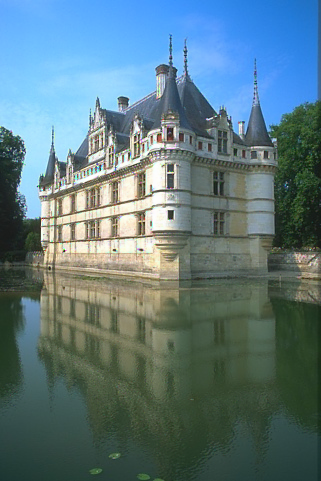}}
\subfigure{\includegraphics[width=0.14\textwidth]{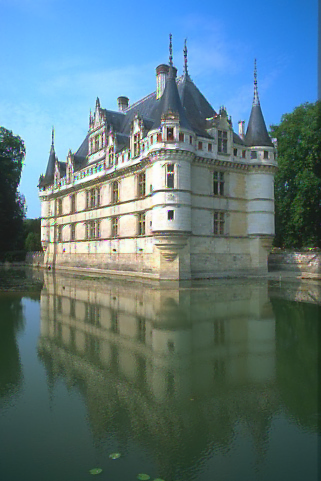}}
\subfigure{\includegraphics[width=0.14\textwidth]{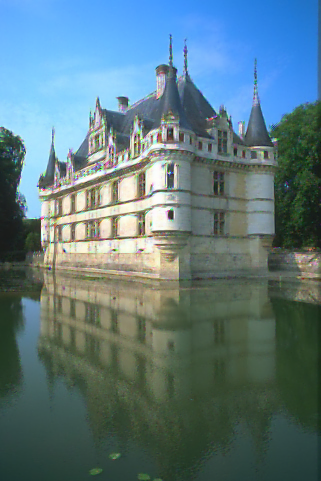}}
\subfigure{\includegraphics[width=0.14\textwidth]{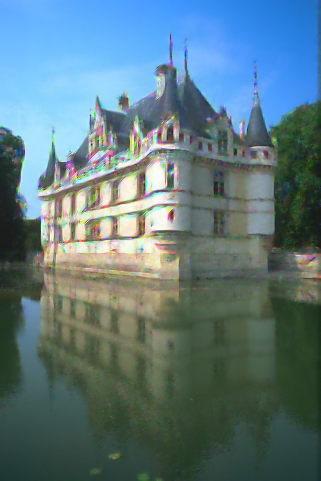}}
\subfigure{\includegraphics[width=0.14\textwidth]{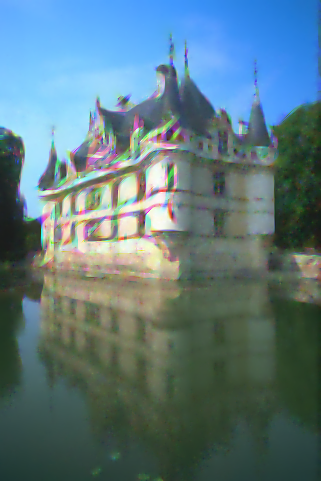}}\\
\vspace{-10pt}
\caption{Denoising results on a real color image from Berkeley database BSDS500. First row: Images (from top to bottom) corrupted by salt-and-pepper noise of $20\%$, $40\%$, $60\%$, $80\%$ and $90\%$; Second row: Media filter results; Third row: TVL1 results; Last row: the proposed TWSO results.}
\label{fig:denoise22}
\vspace{-7pt}
\end{figure}

In addition to demonstrating TWSO for Gaussian noise removal, Figure~\ref{fig:denoise22} and Table~\ref{tb:table2} show the denoising results of images with salt-and-pepper noise by the different methods. Here, we use $p=1$ in (\ref{eq:TWSO}) 
and compare our proposed TWSO method with another two state-of-the-art methods (i.e. media filter and TVL1 \cite{nikolova2004variational}) that have been widely employed in salt-and-pepper noise removal. As seen from the first and second row of Table~\ref{tb:table2}, all three methods perform very well when the image contains low level noise. As the noise increases, the media filter and TVL1 become increasingly worse, introducing more and more artefacts to the resulting images. Both fail in restoring the castle when the image is degraded by 90$\%$ salt-and-pepper nosie. In contrast, TWSO has produced visually more pleasing results against increasing noise and its PSNR ans SSIM remain the highest in all the cases. In fact, the introduction of the tensor in TWSO offers richer neighbourhood information, making it powerful in denoising images corrupted by salt-and-pepper noise.

\begin{table}[h]
\vspace{-5pt}
\caption{Comparison of PSNR and SSIM using different methods on Figure~\ref{fig:denoise22} with different noise densities.}
\vspace{-10pt}
\resizebox{\columnwidth}{!}{
\begin{tabular}{|c||c|c|c|c|c||c|c|c|c|c|}
\hline
                                                        & \multicolumn{5}{c||}{PSNR value}                    & \multicolumn{5}{c|}{SNR value} \\ \hline \hline                         
Noise density                                           & 20$\%$   & 40$\%$    & 60$\%$   & 80$\%$    & 90$\%$     & 20$\%$   & 40$\%$    & 60$\%$   & 80$\%$    & 90$\%$      \\ \hline
Degraded                                                & 11.9259 & 8.92250 & 7.16820 & 5.92240 & 5.41430    & 0.2490     & 0.1304     & 0.0714     & 0.0317    & 0.0164    \\ \hline
Media filter                                            & 24.1623 & 22.6650 & 20.1166 & 16.8871 & 12.0975    & 0.8582     & 0.8146     & 0.7620     & 0.6639    & 0.3265    \\ \hline
TVL1                                                    & 27.9226 & 25.0812 & 23.1954 & 20.8760 & 17.7546    & 0.9405     & 0.8850     & 0.8328     & 0.7544    & 0.6427    \\ \hline
TWSO                                                    & \textbf{59.6130} & \textbf{29.8168} & \textbf{27.7261} & \textbf{25.1963} & \textbf{23.3120}    & \textbf{0.9985}     & \textbf{0.9614}     & \textbf{0.9373}     & \textbf{0.8911}    & \textbf{0.8427}    \\ \hline                                        
\end{tabular}
} 
\label{tb:table2}
\end{table}

\section{Conclusion}
\label{conclusion}
In this paper, we propose the TWSO model for image processing which introduces the anisotropic diffusion tensor to the SOTV model. More specifically, the proposed model is equipped with a novel regulariser for the SOTV model that uses the Frobenius norm of the product of the SOTV Hessian matrix and the anisotropic tensor. The advantage of the TWSO model includes its ability to reduce both the staircase and blurring artefacts in the restored image. To avoid numerically solving the high order PDEs associated with the TWSO model, we develop a fast alternating direction method of multipliers based on a discrete finite difference scheme. Extensive numerical experiments demonstrate that the proposed TWSO model outperforms several state-of-the-art methods for image reconstruction in different cases. Future work will combine the capabilities of the TWSO model for image denoising and inpainting for medical imaging, such as optical coherence tomography \cite{duan2015new}.

\section{Acknowledgement}
We thank Prof. Xue-Cheng Tai of the University of Bergen, Norway for providing the Euler's elastica code for this research. 

\bibliographystyle{unsrt}
\bibliography{Tensor}

\end{document}